\documentclass[runningheads]{llncs}

\usepackage[mobile]{eccv}

\usepackage{eccvabbrv}
\usepackage{graphicx}
\usepackage{booktabs}
\usepackage{wrapfig}
\usepackage{amsmath}
\usepackage{xspace}
\usepackage{xcolor}
\usepackage{textgreek}
\usepackage{bm}
\usepackage{multirow}
\usepackage[accsupp]{axessibility}  
\usepackage[pagebackref,breaklinks,colorlinks,citecolor=eccvblue]{hyperref}
\usepackage[dvipsnames]{xcolor}
\usepackage{orcidlink}
\usepackage{footmisc}
\usepackage{algorithm}
\usepackage{algorithmic}
\usepackage{listings}

\definecolor{mygreen}{RGB}{34,139,34}
\definecolor{mylightblue}{RGB}{0,162,230}
\definecolor{deepyellow}{RGB}{255, 215, 0} 

\newcommand{\model}{\textsc{PixArt-\textSigma{}}\xspace}
\newcommand{\modelalpha}{\textsc{PixArt-$\alpha$}\xspace}
\newcommand{\modelshort}{\textsc{PixArt}\xspace}
\newcommand{\pre}{\textsc{$\alpha$}\xspace}
\newcommand{\now}{\textsc{\textSigma{}}\xspace}

\makeatletter
\def\blfootnote#1{\xdef\@thefnmark{}\@footnotetext{\scriptsize #1}}
\makeatother

\begin{document}

\title{\model: Weak-to-Strong Training of Diffusion Transformer for 4K Text-to-Image Generation}

\author{Junsong Chen\inst{1,2,3*}\and
Chongjian Ge\inst{1,3*} \and
Enze Xie\inst{1*\dagger} \and
Yue Wu\inst{1*}  \and
Lewei Yao\inst{1, 4} \and \\
Xiaozhe Ren\inst{1} \and 
Zhongdao Wang\inst{1} \and 
Ping Luo\inst{3} \and 
Huchuan Lu\inst{2} \and
Zhenguo Li\inst{1}
}
\authorrunning{J. Chen et al.}
\titlerunning{\model: Weak-to-Strong Training of DiT for 4K T2I Generation}
\institute{
$^\text{1 }$Huawei Noah's Ark Lab~~~$^\text{2 }$Dalian University of Technology~~~
$^\text{3 }$HKU~~~$^\text{4 }$HKUST \\
\vspace{0.5em}
Project Page: \url{https://pixart-alpha.github.io/PixArt-sigma-project/}
}

\maketitle

\blfootnote{$*$Equal contribution. Work done during the students' internships at Huawei Noah's Ark Lab. \\$\dagger$Project lead and corresponding author. Thanks to Charan for the 4K dataset collection.}

\begin{abstract}
In this paper, we introduce \model, a Diffusion Transformer model~(DiT) capable of directly generating images at 4K resolution.  \model represents a significant advancement over its predecessor, \modelalpha, offering images of markedly higher fidelity and improved alignment with text prompts. A key feature of \model is its training efficiency. Leveraging the foundational pre-training of \modelalpha, it evolves from the `weaker' baseline to a `stronger' model via incorporating higher quality data, a process we term ``weak-to-strong training''.  The advancements in \model are twofold: (1) High-Quality Training Data: \model incorporates superior-quality image data, paired with more precise and detailed image captions. (2) Efficient Token Compression: we propose a novel attention module within the DiT framework that compresses both keys and values, significantly improving efficiency and facilitating ultra-high-resolution image generation. Thanks to these improvements, \model achieves superior image quality and user prompt adherence capabilities with significantly smaller model size~(0.6B parameters) than existing text-to-image diffusion models, such as SDXL~(2.6B parameters) and SD Cascade~(5.1B parameters).  Moreover, \model's capability to generate 4K images supports the creation of high-resolution posters and wallpapers, efficiently bolstering the production of high-quality visual content in industries such as film and gaming.

\end{abstract}

\section{Introduction}
\label{sec:intro}

\begin{figure}[!ht]
\centering
\includegraphics[width=0.92\linewidth]{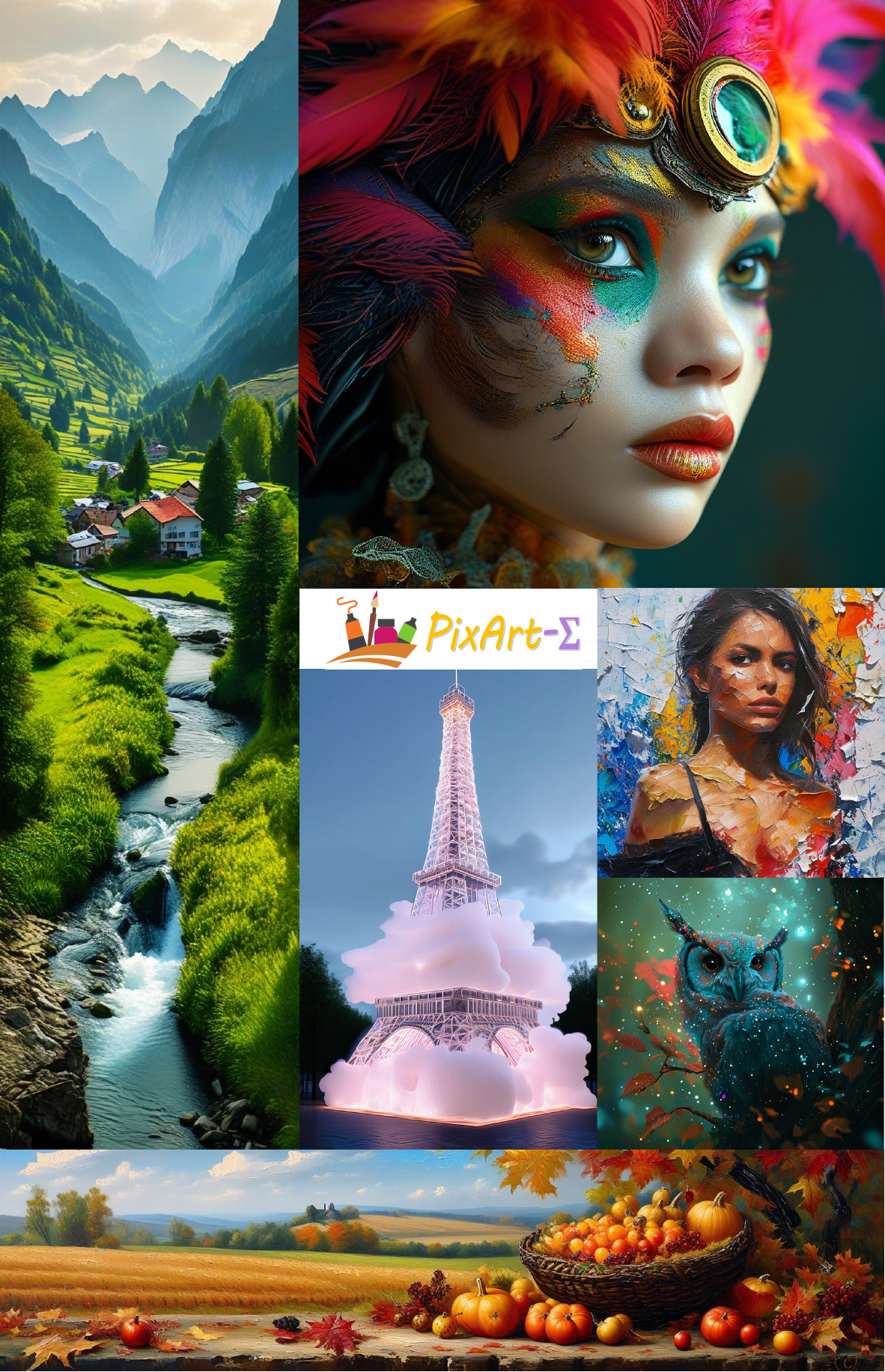}
\vspace{-1em}
\caption{\textbf{\footnotesize Images generated by \model. } The model can output photo-realistic, high aesthetic, extreme aspect ratio, multi-style images, and follow user instructions.}
\label{fig:teaser}
\end{figure}

The recent emergence of high-quality Text-to-Image (T2I) models has profoundly impacted the AI Generated Content (AIGC) community. This includes both proprietary models such as DALL·E 3~\cite{Dalle-3}, Midjourney~\cite{Midjourney}, as well as open-source models like Stable Diffusion~\cite{podell2023sdxl} and \modelalpha~\cite{chen2023pixartalpha}. Nonetheless, developing a top-tier T2I model involves considerable resources; for instance, training SD1.5 from scratch necessitates about 6000 A100 GPU days\cite{podell2023sdxl}, posing a substantial barrier to individual researchers with limited resources and impeding innovation within the AIGC community. Over time, the AIGC community will gain access to continuously updated, higher-quality datasets and more advanced algorithms. A pivotal question is: \textit{how can we efficiently integrate these new elements into an existing model, achieving a more powerful version within the constraints of limited resources?}

To explore this issue, our research focuses on enhancing \modelalpha, an efficient T2I training method. \modelalpha represents an early venture within the DiT framework, a model structure with significant potential, as evidenced by works such as GenTron~\cite{chen2023gentron}, Sora~\cite{Sora} and Stable Diffusion 3~\cite{sd3}. To maximize this potential, we build upon the pre-trained foundation of \modelalpha, integrating advanced elements to facilitate its continuous improvement, resulting in a more powerful model, \model. We refer to this process of evolving from a relatively weaker baseline to a stronger model through efficient training as ``weak-to-strong training''. Specifically, to achieve ``weak-to-strong training'', we introduce the following enhancements:

\textbf{Higher-Quality Training Data}: 
We collect a high-quality dataset superior to that used in \modelalpha, focusing on two key aspects:
(i) \textbf{High-quality images}: The dataset comprises 33M high-resolution images sourced from the Internet, all exceeding 1K resolution, including 2.3M images with resolutions around 4K. These images are predominantly characterized by their high aesthetic and encompass a wide range of artistic styles.
(ii) \textbf{Dense and accurate captions}: To provide more precise and detailed captions for the aforementioned images, we replace the LLaVA~\cite{liu2023llava} used in \modelalpha with a more powerful image captioner, Share-Captioner~\cite{chen2023sharegpt4v}. Furthermore, to improve the model's alignment capacity between the textual and visual concepts, we extend the token length of the text encoder (\ie, Flan-T5~\cite{t5}) to approximately 300 words. We observe these improvements effectively eliminate the model's tendency for hallucination, leading to higher-quality text-image alignment.

\textbf{Efficient Token Compression}: To enhance \modelalpha, we expand its generation resolution from 1K to 4K. Generating images at ultra-high resolutions (\eg, 2K/4K) introduces a significant increase in the number of tokens, leading to a substantial rise in computational demand. To address this challenge, we introduced a self-attention module with key and value token compression tailored to the DiT framework. Specifically, we utilize group convolutions with a stride of 2 for local aggregation of keys and values. Additionally, we employ a specialized weight initialization scheme, allowing for a smooth adaptation from a pre-trained model without KV compression.  This design effectively reduces training and inference time by  $\sim$34\% for high-resolution image generation.

\textbf{Weak-to-Strong Training Strategy}: we propose several fine-tuning techniques to rapidly adapt from a weak model to a strong model efficiently.  
That includes (1) replacing with a more powerful Variational Autoencoder (VAE)~\cite{podell2023sdxl}, (2) scaling from low to high resolution, and (3) evolving from a model without Key-Value (KV) compression to one with KV compression. These outcomes confirm the validity and effectiveness of the  ``weak-to-strong training'' approach.

\begin{figure}[!t]
\centering
\includegraphics[width=0.99\linewidth]{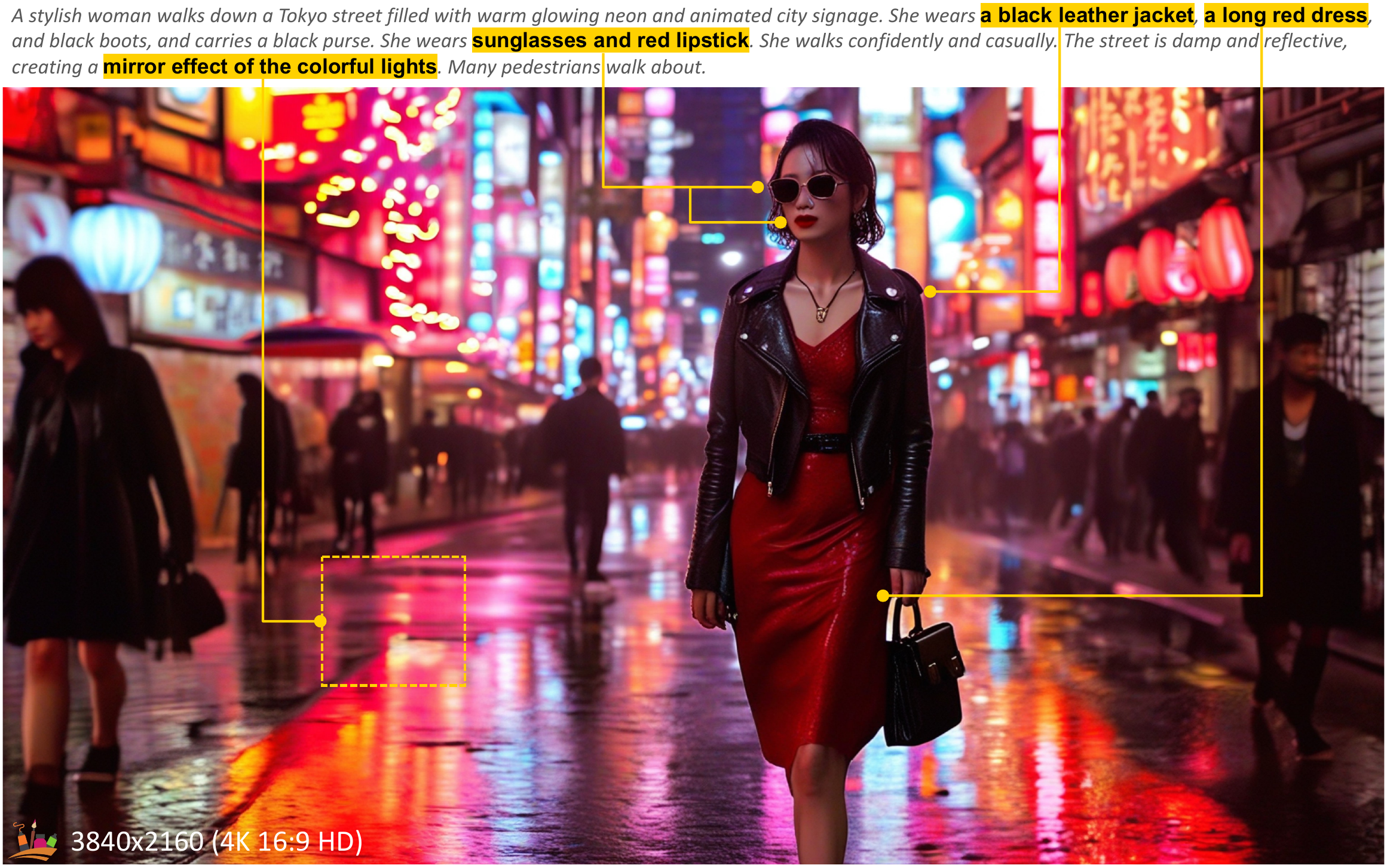}
\vspace{-1em}
\caption{ \textbf{\footnotesize 4K image generation with complex dense instructions.} \model can directly generate 4K resolution images without post-processing, and accurately respond to the given prompt.}
\label{fig:4K}
\end{figure}

\begin{figure}[!ht]
\centering
\includegraphics[width=0.99\linewidth]{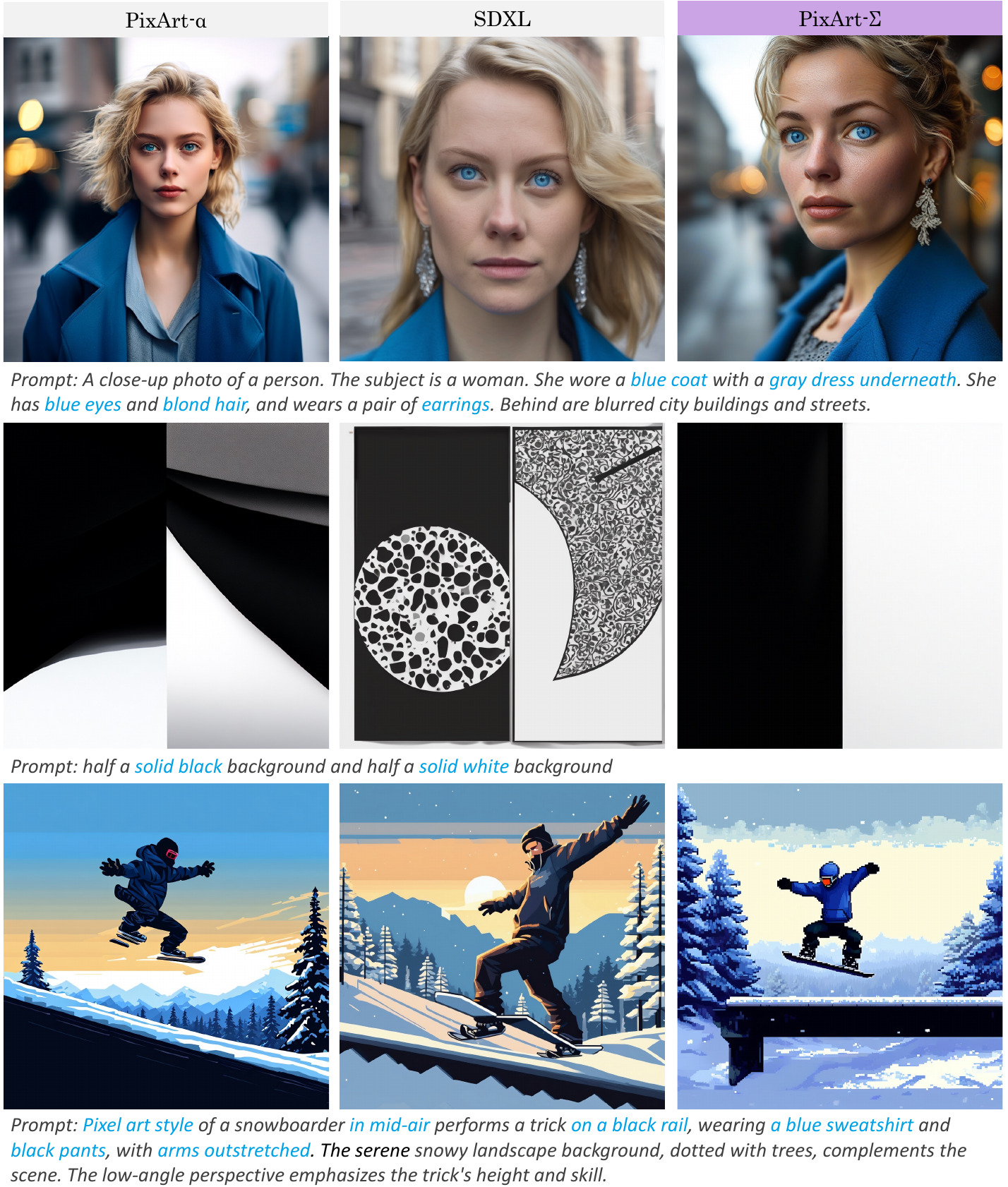}
\vspace{-1em}
\caption{\textbf{\footnotesize Comparison of \model with open-source models, e.g., \modelalpha and SDXL}: Compared with \modelalpha, \model improves the realism of portraits and the capability of semantic analysis. Compared with SDXL, our method has a better ability to follow user instructions. The keywords are highlighted as \textit{\textcolor{mylightblue}{blue}}.}
\label{fig:compare_alpha_sdxl_sigma}
\vspace{-1em}
\end{figure}

\begin{figure}[!t]
\centering
\footnotesize
\includegraphics[width=0.99\linewidth]{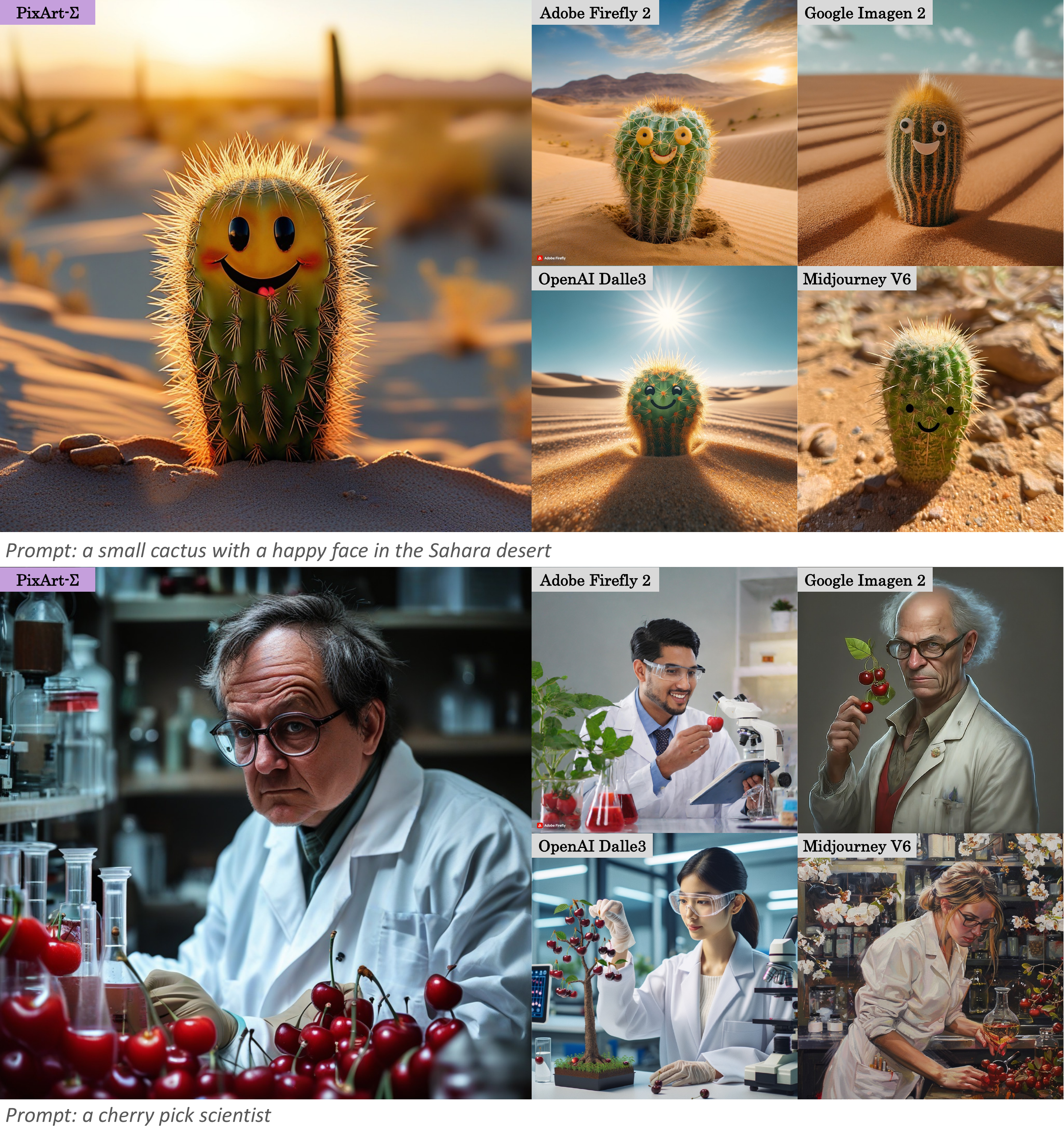}
\vspace{-1em}
\caption{\textbf{Compare \model and four other T2I products}: Firefly 2, Imagen 2, Dalle 3, and Midjourney 6. Images generated by \model are very competitive with these commercial products.}
\label{fig:compare_sota}
\end{figure}

Through the proposed improvements, \model achieves high-quality 4K resolution image generation at a minimal training cost and model parameters. Specifically, fine-tuning from a pre-trained model, we additionally utilize only \textbf{9\%} of the GPU days required by \modelalpha to achieve a strong 1K high-resolution image generation model, which is impressive considering we replace with new training data and a more powerful VAE. Moreover, we use only \textbf{0.6B} parameters while SDXL~\cite{podell2023sdxl} and SD Cascade\cite{pernias2023wurstchen} use 2.6B and 5.1B parameters respectively. Images generated by \model possess an aesthetic quality comparable to current top-tier T2I products, such as DALL·E 3~\cite{Dalle-3} and MJV6~\cite{Midjourney} (as illustrated in Fig.~\ref{fig:compare_sota}). Additionally, \model also demonstrates exceptional capability for fine-grained alignment with textual prompts (as shown in Fig.~\ref{fig:4K} and~\ref{fig:compare_alpha_sdxl_sigma}).

\section{Related Work}

\noindent \textbf{Diffusion Transformers.}
The Transformer architecture has achieved remarkable success across various domains, such as language modeling~\cite{radford2018improving,radford2019language}, computer vision\cite{touvron2021training,liu2021swin,yuan2021tokens,zheng2021rethinking}, and other areas~\cite{von2211self,
chandra2023transformer}. In the realm of diffusion models,  DiT~\cite{Peebles2022DiT} and UViT~\cite{bao2023all} pioneer the use of Transformer architecture. Subsequent works, including DiffiT~\cite{diffit}, SiT~\cite{sit}, and FiT~\cite{fit}, have improved upon DiT’s architecture, while ~\cite{gao2023masked,zheng2023fast} enhance training efficiency through masked modeling techniques. For Text-to-Image (T2I) synthesis, \modelalpha~\cite{chen2023pixartalpha} explore efficient T2I training schemes, achieving the first Transformer-based T2I model capable of generating 1024px high-quality images. GenTron~\cite{chen2023gentron} explores the flexibility and scalability of diffusion Transformers in both image and video generation. The recent advent of the powerful video generation model Sora~\cite{Sora} has further underscored the potential of Diffusion Transformers. In this work, for the first time, we explore using the Transformer architecture to generate 4K ultra-high-resolution images directly, tackling the computational complexity challenges posed by involving long-sequence tokens.

\noindent \textbf{High Resolution Image Generation} greatly enhances visual quality and is important in various industries such as film and gaming. However, increasing image resolution introduces challenges due to the substantial increase in computational demands. Numerous methods have been explored in this direction. For instance, Imagen ~\cite{saharia2022photorealistic}, GigaGAN~\cite{kang2023gigagan} and Stable Diffusion~\cite{rombach2022high} introduce an additional super-resolution network, while Stable Cascade~\cite{pernias2023wurstchen} employs multiple diffusion networks to increase resolution progressively. These combined-model solutions, however, can introduce cumulative errors.
On the other hand, works like SDXL~\cite{podell2023sdxl}, DALL·E 2~\cite{Dalle-2}, Playground~\cite{playground} and \modelalpha~\cite{chen2023pixartalpha} have attempted to generate high-resolution images using diffusion models directly. Nevertheless, these efforts are capped at generating images with resolutions up to 1024px due to computational complexity. In this paper, we push this boundary to 4K resolution, significantly enhancing the visual quality of the generated content.

\noindent \textbf{Efficient Transformer architecture.}
The self-attention mechanism in Transformer suffers from quadratic computational complexity with respect to the number of tokens, which hinders the scaling up of token quantity. Many works have sought improvements in this area:
(1) Sparse Attention\cite{wang2021pyramid,wang2022pvt,xie2021segformer,chen2023sparsevit,chen2021chasing}, which reduces the overall computational load by selectively processing a subset of tokens. For instance, PVT v2~\cite{wang2022pvt} employs a convolutional kernel to condense the space of the key and value, thus lowering the complexity involved in computing the attention.
(2) Local Attention\cite{liu2021swin,zhu2020deformable,xia2022vision,ge2023advancing} focuses attention within nearby regions; notably, Swin Transformer~\cite{liu2021swin} utilizes window-based attention to limit computations to within a specified window size.
(3) Low-rank/Linear Attention~\cite{wang2020linformer,choromanski2020rethinking,lu2021soft}. The Linformer~\cite{wang2020linformer} reduces the computational complexity of the self-attention mechanism through low-rank approximations. In this paper, inspired by PVT v2~\cite{wang2022pvt}, we employ a self-attention mechanism based on key/value compression to mitigate the high complexity of processing 4K images.

\section{Framework}

\subsection{Data Analysis} \label{sec31_data_analysis}
\begin{figure}[!ht]
\vspace{-2em}
\centering
\includegraphics[width=0.99\linewidth]{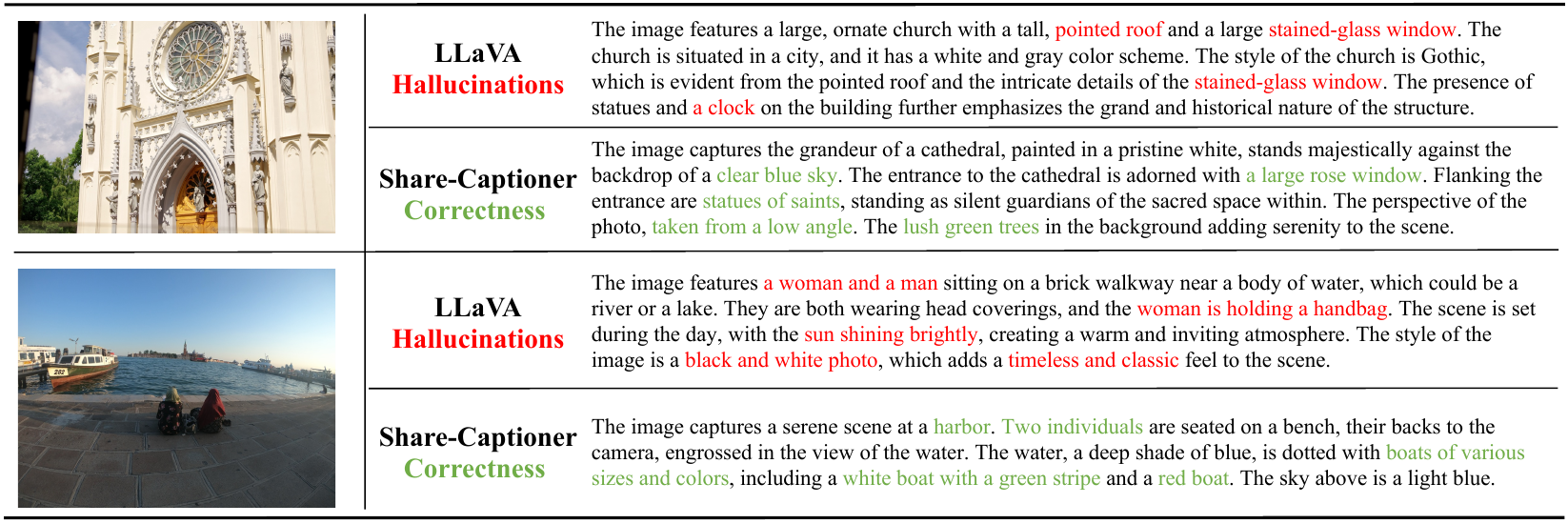}
\vspace{-1em}
\caption{\textbf{\footnotesize Comparative illustration of hallucinations}: Contrasting differences in hallucination occurrences between LLaVA and Share-Captioner, with \textcolor{red}{red} indicating hallucinations and {\color{mygreen} green} denoting correctness.
\vspace{-2em}
}
\label{fig:compare_captions}
\end{figure}

\noindent\textbf{Higher Aesthetic and higher Resolution.}
To enhance the aesthetic quality of our dataset, we expand our internal data from 14M to 33M.  
For clarity, we name the two datasets Internal-\pre and Internal-\now, respectively. 
Note that this expansion still falls short compared to the vast images utilized by currently available open-source models like SD v1.5, which uses 2B data. We demonstrate that effective training strategies with limited data amount can still obtain a strong T2I model.

The images within Internal-\now are above 1K resolution. To facilitate 4K resolution generation, we additionally collect a dataset of 8M real photographic images at 4K resolution. To ensure aesthetic quality, we employ an aesthetic scoring model (AES)~\cite{aes} to filter these 4K images. This process yields a highly refined dataset of 2M ultra-high-resolution and high-quality images.

Interestingly, we have observed that as the resolution of the images increases, there is an improvement in the model's fidelity (Fréchet Inception Distance~(FID)~\cite{heusel2017gans}) and semantic alignment (CLIP Score), which underscores the importance of the capabilities of generating high-resolution images.

\renewcommand{\tabcolsep}{1.5pt}
\def\swfive{0.30\linewidth}

\begin{wrapfigure}{r}{0.5\textwidth}
\centering
\vspace{-8mm}
\includegraphics[width=1\linewidth]{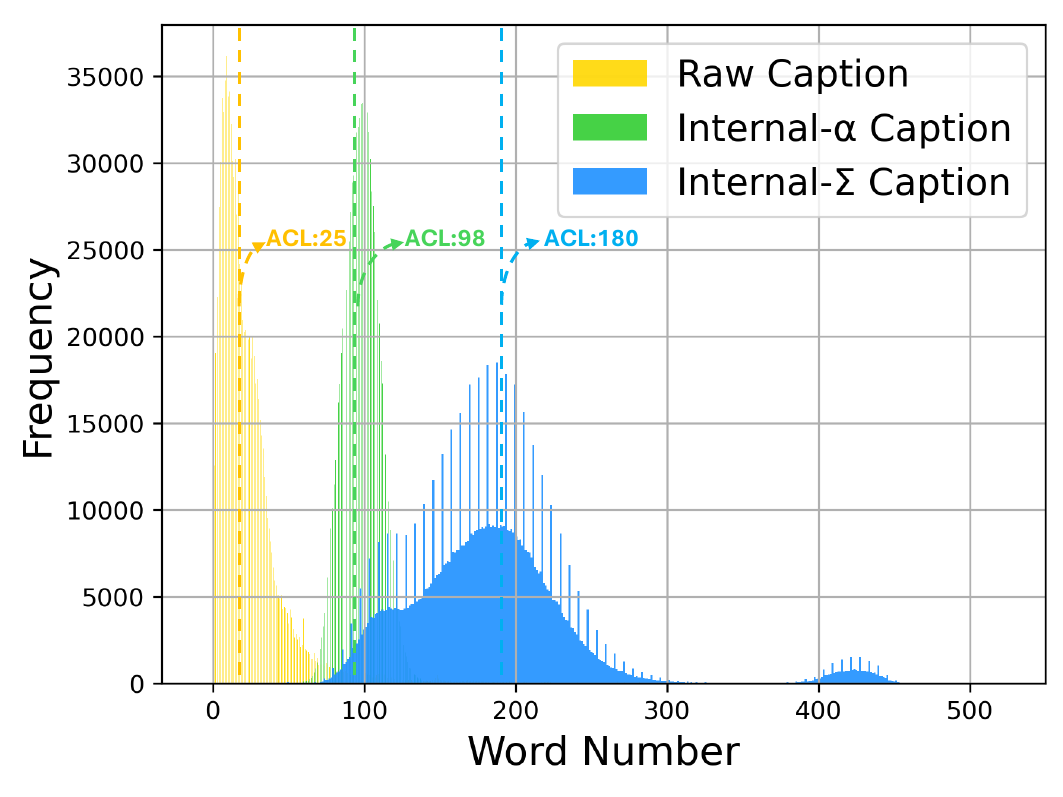}
\vspace{-2em}
\caption{\textbf{ Histogram Visualization of the Caption Length.} We randomly select 1M captions from the \textcolor{deepyellow}{\textbf{raw captions}}, \textcolor{mygreen}{\textbf{Internal-$\alpha$}}, and \textcolor{blue}{\textbf{Internal-\textSigma}} to draw the corresponding histogram. \textit{ACL} denotes the average caption length.}
\label{fig:token_histogram}
\vspace{-2em}
\end{wrapfigure}

\noindent\textbf{Better Text-Image Alignment.} 
Recent works such as \modelalpha~\cite{chen2023pixartalpha} and DALL-E 3~\cite{Dalle-3} emphasize the significance of text-image description alignment. Strengthening this alignment is crucial for boosting model capabilities. To refine our collected ``raw'' descriptions further, we focus on improving both the length and accuracy of our captions. Notably, our captions (Internal-\textSigma{}) show several advantages over the one used in \modelalpha (Internal-$\alpha$) in the following aspects:

\noindent 1. \texttt{Enhanced caption accuracy}: 
As depicted in Fig.~\ref{fig:compare_captions}, LLaVa used in \modelalpha has a certain hallucination problem. We leverage a more powerful Visual-language model, i.e., Share-Captioner~\cite{chen2023sharegpt4v}, to generate detailed and correct captions, augmenting the collected raw prompts. 

\noindent 2. \texttt{Increased caption length}: As shown in Tab.~\ref{tab:laion_sam_llava_compare} and Fig.~\ref{fig:token_histogram}, the average caption length increased significantly to 180 words, highly enhancing the descriptive power of the captions. Additionally, we extend the token processing length of the text encoder from 120 tokens (as in Internal-$\alpha$) to 300 tokens. Our model is trained on a mix of long (Share-Captioner) and short (raw) captions with a ratio of 60\% and 40\%, respectively. This approach enhances the diversity of textual descriptions and mitigates potential biases that might arise from solely relying on generative captions.

Tab. ~\ref{tab:laion_sam_llava_compare} demonstrates a summary for both Internal-\pre and -\now, where we assess the diversity of the datasets through various metrics, including the noun variety, total noun count, average caption length, and average nouns per image.

\begin{table}[t]
\centering
    \vspace{-1em}
    \caption{\textbf{\footnotesize Statistics of noun concepts for different datasets.} {\bf VN}: valid distinct nouns (appearing more than 10 times); {\bf DN}: total distinct nouns; {\bf Average}: average noun count per image; {\bf ACL}: Average Caption length.} 
    \vspace{-1em}
\label{tab:laion_sam_llava_compare}{
\resizebox{0.95\linewidth}{!}{ 
\begin{tabular}{ccccccc}
\toprule
\bf Dataset & \bf Volume & \bf Caption & \bf VN/DN  & \bf Total Noun & \bf ACL & \bf Average \\
\midrule
Internal-\pre & 14M & Raw & 187K/931K& 175M & 25 & 11.7/Img \\
Internal-\pre & 14M & LLaVA & 28K/215K & 536M & 98 & 29.3/Img \\
Internal-\pre & 14M & Share-Captioner & 51K/420K & 815M & 184 & 54.4/Img \\
\midrule
Internal-\now & 33M & Raw & 294K/1512K & 485M & 35 & 14.4/Img \\
Internal-\now & 33M & Share-Captioner & 77K/714K & 1804M & 180 & 53.6/Img \\
4K-\now       & 2.3M & Share-Captioner & 24K/96K & 115M & 163 & 49.5/Img \\
\bottomrule
\end{tabular}
}
}
\vspace{-1em}
\end{table}

\noindent\textbf{High-Quality Evaluation Dataset.}\label{sec:high_evaluate}
Most SoTA T2I models chose MSCOCO~\cite{microsoftcoco} as the evaluation set to assess the FID and CLIP Scores. However, we observe evaluations conducted on the MSCOCO dataset may not adequately reflect a model's capabilities in aesthetics and text-image alignment. Therefore, we propose a curated set comprising 30,000 high-quality, aesthetically pleasing text-image pairs to facilitate the assessment. The selected samples of the dataset are presented in the appendix.
This dataset is designed to provide a more comprehensive evaluation of a model's performance, particularly in capturing the intricacies of aesthetic appeal and the fidelity of alignment between textual descriptions and visual content. Unless otherwise specified, the evaluation experiments in the paper are conducted in the collected High-Quality Evaluation Dataset.

\subsection{Efficient DiT Design}\label{sec:efficient_design}
An efficient DiT network is essential since the computational demand significantly increases when generating images at ultra-high resolutions. 
The attention mechanism plays a pivotal role in the efficacy of Diffusion Transformers, yet its quadratic computational demands significantly limit model scalability, particularly at higher resolutions \eg, 2K and 4K. Inspired by PVT v2~\cite{wang2022pvt}, we incorporate KV compression within the original \modelalpha's framework to address the computational challenges. 
This design adds a mere 0.018\% to the total parameters yet achieves efficient reduction in computational costs via token compression, while still preserving both spatial and semantic information.

\begin{wrapfigure}{r}{0.5\textwidth}
\centering
\vspace{-8mm}
\includegraphics[width=1.0\linewidth]{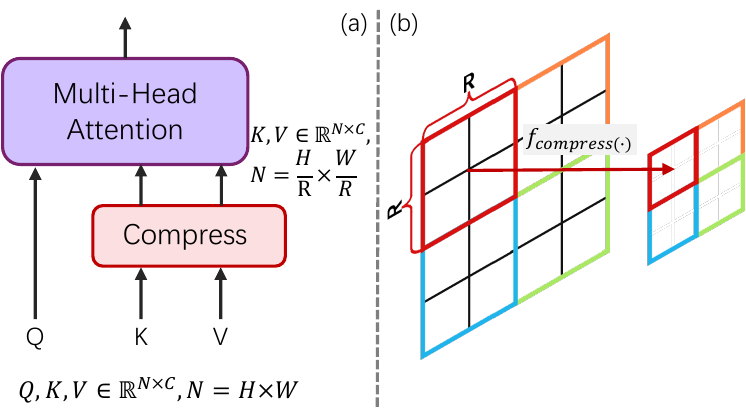}
\vspace{-2em}
\caption{\textbf{Design of KV Token Compression.} We merge KV tokens in spatial space to reduce the computation complexity.}\label{fig:method}
\vspace{-10mm}
\end{wrapfigure}

\noindent \textbf{Key-Value~(KV) Token Compression.}
Our motivation stems from the intriguing observation that applying key-value (KV) token compression directly to the pre-trained \modelalpha can still generate reasonable images. This suggests a redundancy in the features.
Considering the high similarity within adjacent $R \times R$ patches, we assume that feature semantics within a window are redundant and can be compressed reasonably.
We propose KV token compression, which is denoted as $f_{c}(\cdot)$, to compress token features within a $R \times R$ window through a compression operator, as depicted in Fig.~\ref{fig:method}.

Furthermore, to mitigate the potential information loss caused by KV compression in self-attention computation, we opt to retain all the tokens of queries~(Q). This strategy allows us to utilize KV compression effectively while mitigating the risk of losing crucial information. By employing KV compression, we enhance the efficiency of attention computations and reduce the computation complexity from \(O(N^2)\) to \(O\left(\frac{N^2}{R^2}\right)\), thereby making the computational cost of directly generating high-resolution images manageable.

\begin{equation}
    \text{Attention}(Q, K, V) = \text{softmax}\left(\frac{Q \cdot f_{c}(K)^T}{\sqrt{d_k}}\right)f_{c}(V)
\end{equation}

We compress deep layers using the convolution operator ``Conv2$\times2$'' with specific initialization. 
Detailed experiments on other design variants are discussed in Sec.~\ref{sec:ablation}. 
Specifically, we design a specialized convolution kernel initialization ``Conv Avg Init'' that utilizes group convolution and initializes the weights $w = \frac{1}{R^2}$, equivalent to an average operator. This initialization strategy can initially produce coarse results, accelerating the fine-tuning process while only introducing 0.018\% additional parameters.

\subsection{Weak-to-Strong Training Strategy}
We propose several efficient training strategies to enhance the transition from a ``weak'' model to a ``strong'' model. These strategies encompass VAE rapid adaptation, high-resolution fine-tuning, and KV Token compression.

\begin{figure}[t]
\centering
\vspace{-1em}
\includegraphics[width=0.99\linewidth]{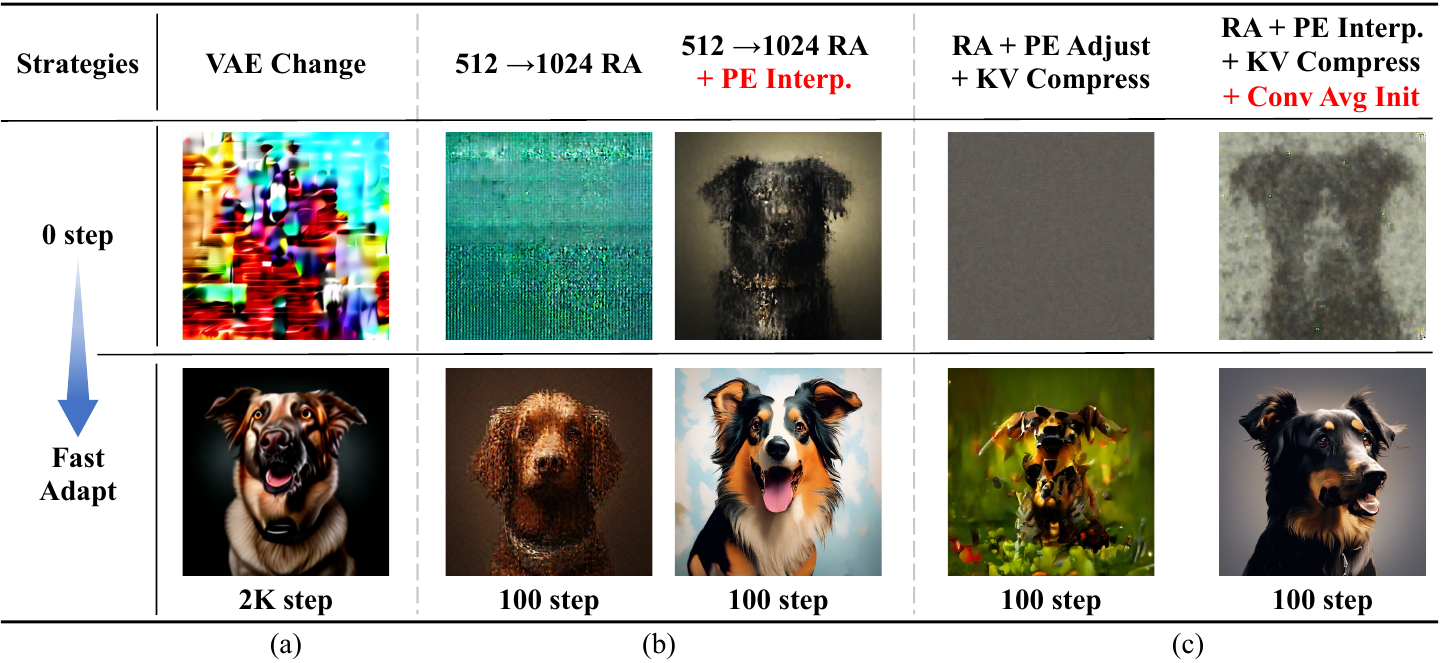}
\vspace{-1em}
\caption{\footnotesize This illustration demonstrates how our training strategy designs expedite the model's convergence during the transition to VAEs, adjustment to higher resolutions, and the KV compression process, facilitating rapid learning from weak to strong.
}
\vspace{-2em}
\label{fig:train_strategy}
\end{figure}
\begin{wraptable}{r}{0.5\textwidth}
\centering
    \vspace{-12mm}
    \caption{We fine-tune a high-resolution model from a low-resolution model and observe that even fine-tuning for a relatively short duration, such as 1K steps, can still yield high-quality results.} 
    \vspace{2mm}
\label{tab:finetune_highres}{
\resizebox{0.8\linewidth}{!}{ 
\begin{tabular}{lccc}
\toprule
Resolution & Iterations & FID $\downarrow$ &  CLIP $\uparrow$ \\
\midrule
256 & 20K & 16.56 & 0.270 \\
256 $\rightarrow$ 512 & 1K & 9.75 & 0.272 \\
256 $\rightarrow$ 512 & 100K & 8.91 & 0.276 \\
\bottomrule
\end{tabular}
}
}
\vspace{-8mm}
\end{wraptable}

\noindent\textbf{Adapting model to new VAEs.}
As VAEs continue to develop, training T2I models from scratch is resource-intensive. We replace \modelalpha's VAE with SDXL's VAE and continue fine-tuning the diffusion model. We observe a rapid convergence phenomenon that fine-tuning quickly converges at 2K training steps as shown in Fig~\ref{fig:train_strategy}~(a). Fine-tuning is more efficient when dealing with VAE model transferring and negates the necessity of training from scratch.

\noindent\textbf{Adapting to Higher-Resolution.}
When we fine-tune from a low-resolution (LR) model to a high-resolution (HR) model, we observe a performance degradation as shown in Fig.~\ref{fig:train_strategy}~(b), which we attribute to discrepancies in positional embeddings (PE) between different resolutions. To mitigate this issue, we utilize the ``PE Interpolation'' trick~\cite{xie2023difffit,chen2023pixartalpha}: initializing the HR model's PE by interpolating the LR model's PE, significantly enhancing the HR model's initial status and expediting the fine-tuning process. We can obtain visually pleasing images even within only 100 training iterations.
Besides, we quantitatively evaluate the model's performance change as illustrated in Tab.~\ref{tab:finetune_highres}. The fine-tuning quickly converges at 1K steps, and further training slightly improves the performance.
This illustrates that using the ``PE Interpolation'' trick enables rapid convergence of higher resolution generation, obviating the need for training from scratch for generating at higher resolutions.

\noindent\textbf{Adapting model to KV compression.} 
We can use KV compression directly when fine-tuning from LR pre-trained models without KV compression. 
As shown in Fig.~\ref{fig:train_strategy}~(c), with our ``Conv Avg Init.'' strategy, \model starts from a better initial state, making converging easier and faster. Notably, \model performs satisfied visual results even within 100 training steps.
Finally, through the KV compression operators and compression layers design in Sec~\ref{sec:efficient_design}, we can reduce $\sim$34\% of the training and inference time.

\section{Experiment}

\subsection{Implementation Details}
\textbf{Training Details.} We follow Imagen~\cite{saharia2022photorealistic} and \modelalpha~\cite{chen2023pixartalpha} to employ the T5~\cite{t5}'s encoder~(\ie, Flan-T5-XXL) as the text encoder for conditional feature extraction, and use \modelalpha~\cite{chen2023pixartalpha} as our base diffusion model. 
Unlike most works that extract fixed 77 text tokens, we adjust the length of text tokens from \modelalpha's 120 to 300, as the caption curated in Internal-\textSigma{} is much denser to provide highly fine-grained details. To capture the latent features of input images, we employ a pre-trained and frozen VAE from SDXL~\cite{podell2023sdxl}. Other implementation details are the same as \modelalpha. Models are finetuned on the \modelalpha's 256px pre-trained checkpoint with the position embedding interpolation trick~\cite{chen2023pixartalpha}. 
Our final models, including 1K resolution, are trained on 32 V100 GPUs.
We additionally use 16 A100 GPUs to train the 2K and 4K image generation models. For further information, please refer to the appendix.

Note that we use CAME optimizer~\cite{luo2023came} with a weight decay of 0 and a constant learning rate of 2e-5, instead of the regular AdamW~\cite{loshchilov2017adamw} optimizer. This helps us reduce the dimension of the optimizer's state, leading to lower GPU memory without performance degradation.

\noindent\textbf{Evaluation Metrics.} To better illustrate aesthetics and semantic ability, we collect 30K high-quality text-image pairs (as mentioned in Sec.~\ref{sec:high_evaluate}) to benchmark the most powerful T2I models. We mainly evaluate \model via human and AI preference study since FID~\cite{rombach2022high} metrics may not adequately reflect the generation quality. 
However, we still provide the FID results on the collected dataset in the appendix.

\subsection{Performance Comparisons}
\noindent \textbf{Image Quality Assessment.}
We qualitatively evaluated our methodology against both closed-source text-to-image (T2I) products and open-source models. As illustrated in Fig.~\ref{fig:teaser}, our model can produce high-quality, photo-realistic images with intricate details over diverse aspect ratios and styles. This capability underscores the superior performance of our approach in generating visually compelling content from textual descriptions. As shown in Fig.~\ref{fig:compare_alpha_sdxl_sigma}, we compare \model with open-source models SDXL~\cite{podell2023sdxl} and \modelalpha~\cite{chen2023pixartalpha}, our method enhances the realism of portraits and boosts the capacity for semantic analysis. In contrast to SDXL, our approach demonstrates superior proficiency in adhering to user instructions. 

Not only superior to open-source models, but our method is also very competitive with current T2I closed-source products, as depicted in Fig.~\ref{fig:compare_sota}. \model produces photo-realistic results and adheres closely to user instructions, which is on par with contemporary commercial products.

 \begin{wrapfigure}{r}{0.6\textwidth}
    \centering
    \vspace{-9mm}
    \includegraphics[width=0.5\textwidth]{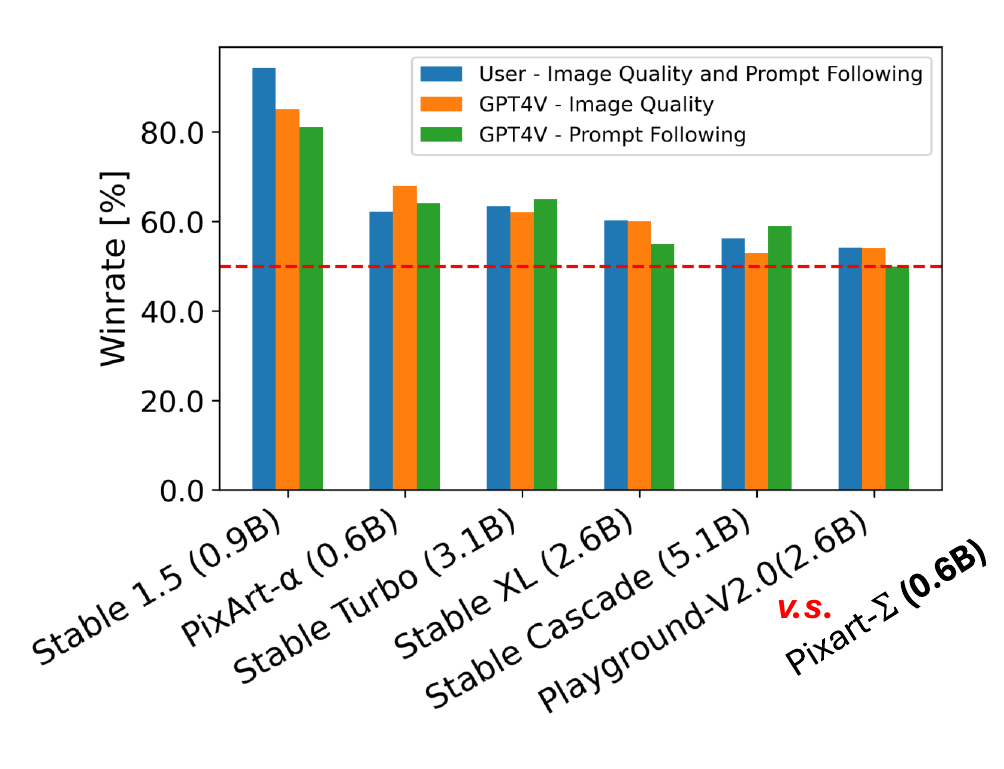}
    \vspace{-6mm}
    \caption{
    \textbf{Human(\textcolor{blue}{blue})/AI(\textcolor{orange}{orange} and \textcolor{mygreen}{green}) preference evaluation against currrent open T2I models.} \model compares favorably against current state-of-the-art T2I models in both image quality and prompt-following.}
    \label{fig:user_study}
    \vspace{-8mm}
\end{wrapfigure}

\noindent \textbf{High-resolution Generation.}
Our method is capable of directly generating 4K resolution images without the need for any post-processing. Additionally, it excels in accurately following complex, detailed, and long text provided by users, as demonstrated in Fig.~\ref{fig:4K}. Thus, users do not need prompt engineering to achieve satisfactory results.

Our approach enables direct 4K image generation. In parallel, studies~\cite{du2023demofusion,he2023scalecrafter} have introduced tuning-free post-processing techniques aimed at generating HR images from LR models or employing super-resolution models~\cite{xue2023raphael} to produce HR images. However, their corresponding results often exhibit artifacts for two primary reasons: (1) Accumulative error may arise due to the cascade pipeline. (2) These methods do not capture the true distribution of 4K images nor learn the alignment between text and 4K images. 
We argue that our method might be a more promising way to generate high-resolution images. Our method yields superior results, and more visual comparison is included in the supplement.

\noindent \textbf{Human/AI~(GPT4V) Preference Study.}
We evaluate the well-trained model in both the human and AI preference study using a subset of 300 captions randomly collected from the High-Quality Evaluation Dataset mentioned in Sec.~\ref{sec31_data_analysis}. We collect images generated by overall six open-source models, including \modelalpha, \model, SD1.5~\cite{rombach2022high}, Stable Turbo~\cite{sauer2023adversarial}, Stable XL~\cite{podell2023sdxl}, Stable Cascade~\cite{pernias2023wurstchen} and Playground-V2.0~\cite{playground-v2}.
We develop a website for the human preference study to display the prompts and their corresponding images. This website was distributed to trained evaluators, who were asked to assess the images, ranking them according to quality and how well they matched the text prompts. The results, illustrated by the blue bar in Fig.~\ref{fig:user_study}, indicate a marked preference for \model over the other six T2I generators. \model generates superior high-quality images that closely follow user prompts, using a much smaller size~(0.6B parameters) compared to existing T2I diffusion models like SDXL~(2.6B parameters) and SD Cascade~(5.1B parameters).

Additionally, in our AI preference study, we employ the advanced multimodal model, GPT-4 Vision~\cite{2023GPT4VisionSC}, as the evaluator. For each trial, we supply GPT-4 Vision with two images: one from \model and another from a competing T2I model. We craft distinct prompts guiding GPT-4 Vision to vote based on image quality and image-and-text alignment. The results, represented by orange and green bars in Fig.~\ref{fig:user_study}, demonstrate consistent outcomes in both human and AI preference studies. Specifically, \model surpasses the baseline, \modelalpha, in effectiveness. Compared to contemporary advanced models such as Stable Cascaded, \model exhibits competitive or superior performance in terms of image quality and instruction-following abilities.

\begin{table*}[t]
\centering
\begin{subtable}[t]{0.42\textwidth}
\begin{center}
    {\resizebox{0.8\textwidth}{!}{
    \begin{tabular}{l c c}
    \toprule
    Layers     & FID $\downarrow$ & CLIP-Score $\uparrow$ \\
    \midrule
        N/A & 8.244 & 0.276 \\
        Shallow (1-14) & 9.278 & 0.275 \\
        Middle (7-20) & 9.063  & 0.276 \\
        Deep (14-27) & 8.532 & 0.275 \\
    \bottomrule
    \end{tabular}
    }
    \caption{Compression layers.}
    \label{tab:ablations-pos}
    }
\end{center}
\end{subtable}
\begin{subtable}[t]{0.44\textwidth}
\begin{center}
    {\resizebox{0.8\textwidth}{!}{
    \begin{tabular}{l c c c}
    \toprule
    Operator     & FID $\downarrow$ & CLIP-Score $\uparrow$ \\
    \midrule
        N/A & 8.244 & 0.276 \\
        Token Discarding & 8.918 & 0.275 \\
        Token Pooling & 9.415 & 0.275 \\
        Conv2$\times$2 & 8.505 & 0.274 \\
    \bottomrule
    \end{tabular}
    }
    \caption{Compression operators.}
    \label{tab:ablations-operator}
    }
\end{center}
\end{subtable}
\begin{subtable}[t]{0.48\textwidth}
\begin{center}
    {\resizebox{0.9\textwidth}{!}{
    \begin{tabular}{l c c c l}
    \toprule
    Res. & Ratio     & FID $\downarrow$ & CLIP-Score $\uparrow$ & Train Latency $\downarrow$ \\
    \midrule
        512 & 1 & 8.244 & 0.276 & 2.3 \\
        512 & 2 &  9.063 & 0.276 & 2.2 \textcolor{red}{(-4\%)}\\
        512 & 4 & 9.606 & 0.276 & 2.1 \textcolor{red}{(-9\%)}\\
        \midrule
        1024 & 1 & 5.685 & 0.277 &  27.5  \\
        1024 & 2 & 5.512 & 0.273 &  22.5 \textcolor{red}{(-18\%)} \\
        1024 & 4 & 5.644 & 0.276 &  20.0 \textcolor{red}{(-27\%)} \\
        1024 & 9 & 5.712 & 0.275 &  17.8 \textcolor{red}{(-35\%)} \\
    \bottomrule
    \end{tabular}
    }
    \caption{Compression rations on different resolutions.}
    \label{tab:ablations-res}
    }
\end{center}
\end{subtable}
\begin{subtable}[t]{0.45\textwidth}
\begin{center}
    {\resizebox{0.9\textwidth}{!}{
    \begin{tabular}{c c l l}
    \toprule
    \multirow{2}{*}{Res.} & \multirow{2}{*}{Ratio} & Train Latency $\downarrow$ & Test Latency  $\downarrow$ \\
         &       &(s/Iter@32BS) & (s/Img)\\
    \midrule
        2K & 1 & 56    & 58 \\
        2K & 4 & 37 \textcolor{red}{(-34\%)} & 38 \textcolor{red}{(-34\%)}  \\
        \midrule
        4K & 1 & 191 & 91  \\
        4K & 4 & 125 \textcolor{red}{(-35\%)} & 60 \textcolor{red}{(-34\%)} \\
    \bottomrule
    \end{tabular}
    }
    \caption{Speed of different resolutions.}
    \label{tab:ablations-speed}
    }
\end{center}
\end{subtable}
\vspace{-1em}
\caption{\textbf{KV-Token Compression Settings in Image Generation.} This study employs FID, CMMD, and CLIP-Score metrics to assess the impact of various token compression components, such as compression ratio, positions, operators, and varying resolutions. Speed calculation in Tab.~\ref{tab:ablations-res} is Second/Iteration/384 Batch-size.
}
\label{tab:ablation-compression}
\vspace{-2em}
\end{table*}

\section{Ablation Studies}\label{sec:ablation}
We conduct ablation studies on generation performance on various KV compression designs. Unless specified, the experiments are conducted on 512px generation. The detailed settings of each ablation experiment are included in the appendix.

\subsection{Experimental settings} 
We use the test set described in Sec.~\ref{sec:high_evaluate} for evaluation.
We employ FID to compute the distributional difference between the collected and generated data for comparative metrics. Furthermore, we utilize CLIP-Score to assess the alignment between prompts and the generated images.

\subsection{Compression Designs}
\noindent \textbf{Compression positions.} 
We implemented KV compression at different depths within the Transformer structure: in the shallow layers (1$\sim$14), the intermediate layers (7$\sim$20), and the deep layers (14$\sim$27). 
As indicated in Tab.~\ref{tab:ablations-pos}, employing KV compression on deep layers notably achieves superior performance. 
We speculate this is because shallow layers typically encode detailed texture content, while deep layers abstract high-level semantic content. Because compression tends to affect image quality rather than semantic information, compressing deep layers can achieve the least loss of information, making it a practical choice for accelerating training but not compromising generation quality.

\noindent \textbf{Compression operators.} We explored the impact of different compression operators. 
We employed three techniques, random discarding, average pooling, and parametric convolution, to compress 2$\times$2 tokens into a single token. As illustrated in Table~\ref{tab:ablations-operator}, the ``Conv 2$\times$2'' method outperforms the others, underscoring the advantage of using a learnable kernel to more effectively reduce redundant features than simple discarding methods.

\noindent \textbf{Compression ratios on different resolutions.} 
We investigated the influence of varying compression ratios on different resolutions. 
As shown in Tab.~\ref{tab:ablations-res}, remarkably, we find that token compression does not affect the alignment between textual and generated images (CLIP Score) but influences the image quality (FID) across resolutions. Although there is a slight degradation in image quality with increasing compression ratios, our strategy brings a training speedup of 18\% to 35\%. This suggests that our proposed KV compression is both effective and efficient for achieving high-resolution T2I generation.

\noindent \textbf{Speed comparisons on different resolutions.}
We further comprehensively validate the speed acceleration in both training and inference in Tab.~\ref{tab:ablations-speed}. Our method can speed up training and inference by approximately 35\% in the 4K generation. Notably, we observe that the training acceleration increases as the resolution rises. For example, the training gradually accelerates from 18\% to 35\% as the resolution increases from 1K to 4K. 
This indicates the effectiveness of our method with increasing resolution, demonstrating its potential applicability to even higher-resolution image generation tasks.
\section{Conclusion}
In this paper, we introduce \model, a Text-to-Image (T2I) diffusion model capable of directly generating high-quality images at 4K resolution. Building upon the pre-trained foundation of \modelalpha, \model achieves efficient training through a novel ``weak-to-strong training'' methodology. This approach is characterized by the incorporation of higher-quality data and the integration of efficient token compression. \model excels at producing high-fidelity images while adhering closely to textual prompts, surpassing the high standards set by its predecessor, \modelalpha. We believe that the innovations presented in \model will not only contribute to advancements in the AIGC community but also pave the way for entities to access more efficient, and high-quality generative models.

\noindent \textbf{Acknowledgement.} We would like to thank Zeqiang Lai, Fei Chen, and Shuchen Xue for discussing and helping implement the DMD part.

\clearpage 
\appendix
\vspace{-8mm}
\section{Appendix}

\subsection{Extension: Inference Acceleration~(\modelshort + DMD).}
\begin{wrapfigure}{l}{0.5\textwidth}
\centering
\vspace{-2mm}
\includegraphics[width=0.5\textwidth]{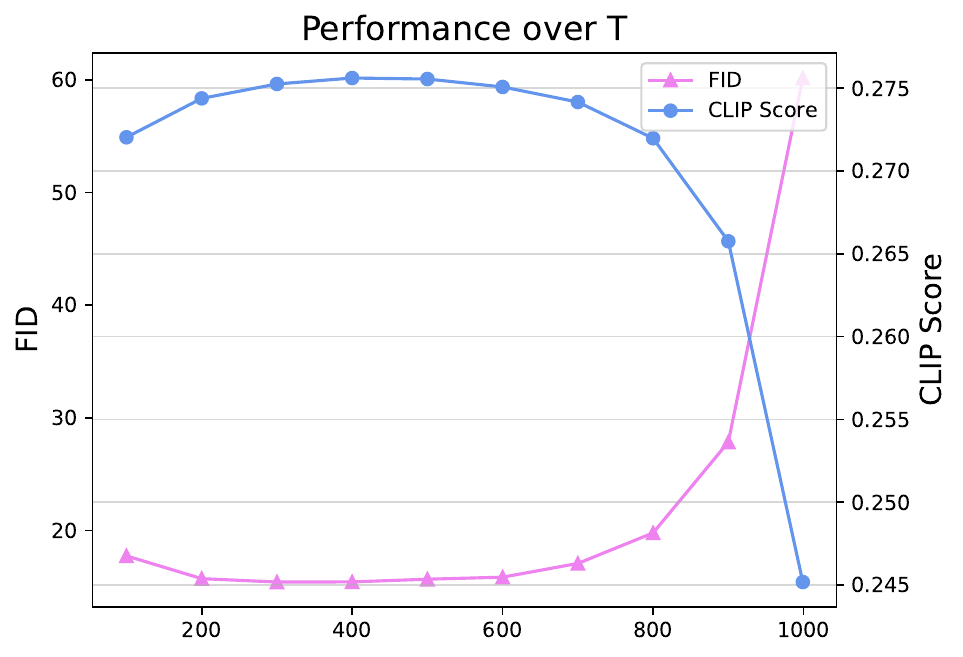}
\vspace{-2em}
\caption{Base model + DMD performance over $\overline{T}$.}
\label{fig:dmd_curve}
\end{wrapfigure}

\begin{figure}[ht]
    \centering
    \includegraphics[width=0.99\textwidth]{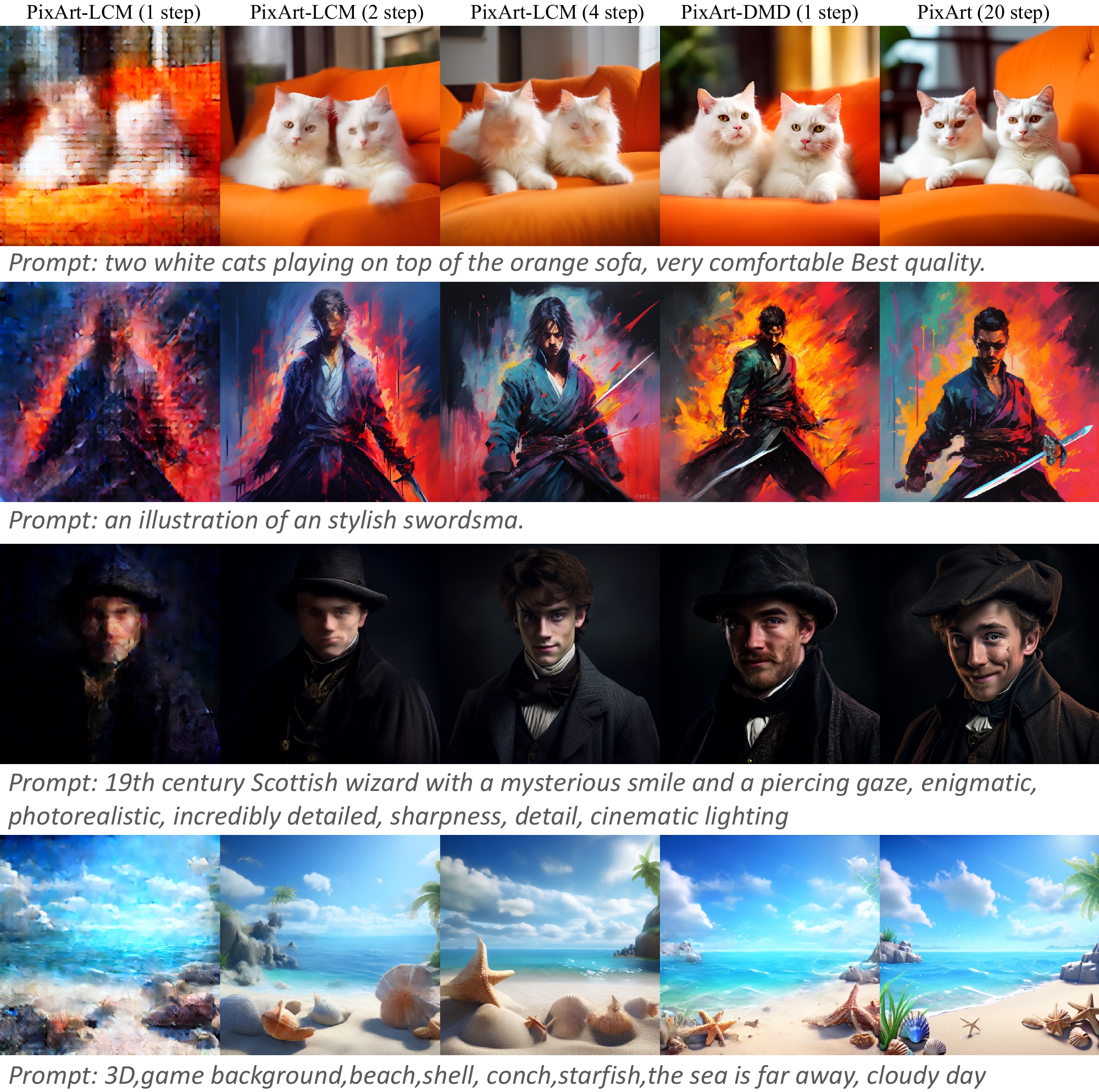}
    \caption{Visual comparison of Base model + DMD performance \vs Base model + LCM model.}   
    \label{fig:dmd_vis}
\vspace{-1mm}
\end{figure}

\noindent\textbf{Experiments.}
To expedite the inference, we integrate \modelshort with DMD~\cite{yin2024onestep}, a one-step inference technique achieved through distribution matching distillation. 
We train a one-step Generator $G_\theta$, and the generated image is denoted as $x_0 = G_\theta(\overline{T}, \rm{text})$.
Initially, we set $\overline{T}$ to the same value as the denoising timesteps during training, that is $\overline{T}=999$ following~\cite{yin2024onestep}. However, we observed 
undesired results and investigated the most suitable value for $\overline{T}$ as shown in Fig.~\ref{fig:dmd_curve}.
Surprisingly, the optimal $\overline{T}$ was found to be 400 rather than 999. This deviation arises from the fact that a smaller $T$ enhances the model's confidence in predicting the noise of $G_\theta$. However, this principle is effective only within a certain range. If $T$ becomes too small, the scenario significantly deviates from the training setting of the base model. Therefore, there exists a trade-off in selecting the $T$ value.
Finally, we compare our method quantitatively as well as qualitatively with \modelshort + LCM~\cite{chen2024pixartdelta} in Tab.~\ref{tab:dmd} and Fig.~\ref{fig:dmd_vis}. We also provide a demo video (attached as \textit{dmd\_demo.mp4}) to compare the speed of \modelshort + DMD with the teacher model.

\renewcommand{\tabcolsep}{1.5pt}
\def\swfive{0.20\linewidth}

\renewcommand{\tabcolsep}{1.5pt}
\def\swfive{0.30\linewidth}

\begin{wraptable}{r}{0.48\textwidth}
\vspace{-20.5em}
\centering
\vspace{-1em}
\caption{\textbf{Comparison of \modelshort + DMD performance compared to \modelshort + LCM.} These experiments are conducted on 512x512 resolution with a batch size of 1.} 
\label{tab:dmd}{
\resizebox{1.0\linewidth}{!}{ 
\begin{tabular}{cccc}
\toprule
\bf Method & FID$\downarrow$ & CLIP$\uparrow$ & Speed$\downarrow$ \\
\midrule
\modelshort + LCM (1 step) &   108.66 &  0.2247 & 0.11s \\
\modelshort + LCM (2 step) &   17.95 &  0.2736 & 0.16s \\
\modelshort + LCM (4 step) &   13.06 &  0.2797 & 0.26s \\
\modelshort + DMD (1 step) &   13.35 & 0.2788 & 0.11s  \\
\midrule
Teacher model (20 steps) &   9.273 & 0.2863 & 1.44s \\
\bottomrule 
\end{tabular}
}
}
\vspace{-1em}
\end{wraptable}

\subsection{Training Details of \model}
In Tab.~\ref{tab:additional_implementation}, we provide detailed information on each training stage of \model, including image resolution, the total volume of training samples, the number of training steps, batch size, learning rate, and the computing time measured in GPU days. Utilizing the Internal-\now dataset and integrating a more advanced VAE, our proposed method quickly adapts to the new VAE, requiring a mere 5 V100 GPU days. Subsequently, we achieve remarkable text-image alignment with just 50 V100 GPU days, showcasing a substantial improvement over \modelalpha at minimal additional training expense, thanks to the proposed "weak-to-strong" strategy, which proves highly efficient.

Notably, applying KV token compression stands out as a significant efficiency booster, markedly reducing the training duration.
For example, when fine-tuning from 512px to 1024px and incorporating KV compression, the training time reduces dramatically from 50 V100 GPU days to a mere 20 V100 GPU days. Likewise, for resolutions of 2K and 4K, the training time diminishes from 20 to 14 A800 GPU days and from 25 to 20 A800 GPU days, respectively. This underscores the remarkable efficacy of KV token compression in augmenting training efficiency.

\begin{table}[ht]
    \centering
    \vspace{-2mm}
    \caption{We report detailed information about each training stage of \model. Note that Internal-\now dataset here includes 33M internal data. The count of GPU days excludes the time for VAE feature extraction and T5 text feature extraction, as we offline prepare both features in advance so that they are not part of the training process and contribute no extra time to it.}
    \label{tab:additional_implementation}
    \resizebox{1.\linewidth}{!}{ 
    \begin{tabular}{l|cccccc}
    \toprule
    Stage & Image Resolution & $\mathrm{\#}$Images & Training Steps & Batch Size & Learning Rate & GPU days \\
    \midrule
    VAE adaption     & 256$\times$256  & 33M   & 8K   & 64$\times$16     & 2$\times{10^{-5}}$ & 5 V100 \\
    Better Text-Image align & 256$\times$256  & 33M   & 80K   & 64$\times$16   & 2$\times{10^{-5}}$ & 50 V100 \\
    \midrule
    Higher aesthetics   & 512$\times$512  & 18M   & 10K  & 32$\times$32 & 2$\times{10^{-5}}$ & 30 V100 \\
    Higher aesthetics   & 1024$\times$1024& 18M   & 5K    & 12$\times$32 & 1$\times{10^{-5}}$ & 50 V100 \\
    KV token compression & 1024$\times$1024& 18M   & 5K    & 12$\times$16 & 1$\times{10^{-5}}$ & 20 V100 \\
    \midrule
    Higher aesthetics   & 2K$\times$2K    & 300K   & 4K    & 4$\times$8 & 2$\times{10^{-5}}$ & 20 A800 \\
    KV token compression & 2K$\times$2K    & 300K   & 4K    & 4$\times$8 & 2$\times{10^{-5}}$ & 14 A800 \\
    Higher aesthetics   & 4K$\times$4K    & 100K   & 2K & 4$\times$8 & 2$\times{10^{-5}}$ & 25 A800 \\
    KV token compression & 4K$\times$4K    & 100K     & 2K & 4$\times$8 & 2$\times{10^{-5}}$ & 20 A800 \\   
    \bottomrule
    \end{tabular}
    }
\vspace{-8mm}
\end{table}

\subsection{Detailed Settings of the Ablation Studies}

In this subsection, we describe the additional experimental setup for the ablation studies. For each study, we evaluated the Fréchet Inception Distance (FID) by comparing images generated by our \model using a set of 30,000 curated prompts against those from the High-Quality Evaluation Dataset. Additionally, we assessed the Clip-Scores by comparing the model-generated images to the original 30,000 prompts. Our findings suggest that the number of images used for testing can influence the FID scores; for instance, a dataset comprising 10,000 images typically yields higher FID scores. Besides, for the conducted experiments on compression positions and operators, we test the respective models with images generated at a resolution of 512px.

\subsection{Samples of High-Quality Evaluation Dataset}
\begin{figure}[!ht]
\vspace{-1em}
\centering
\includegraphics[width=0.99\linewidth]{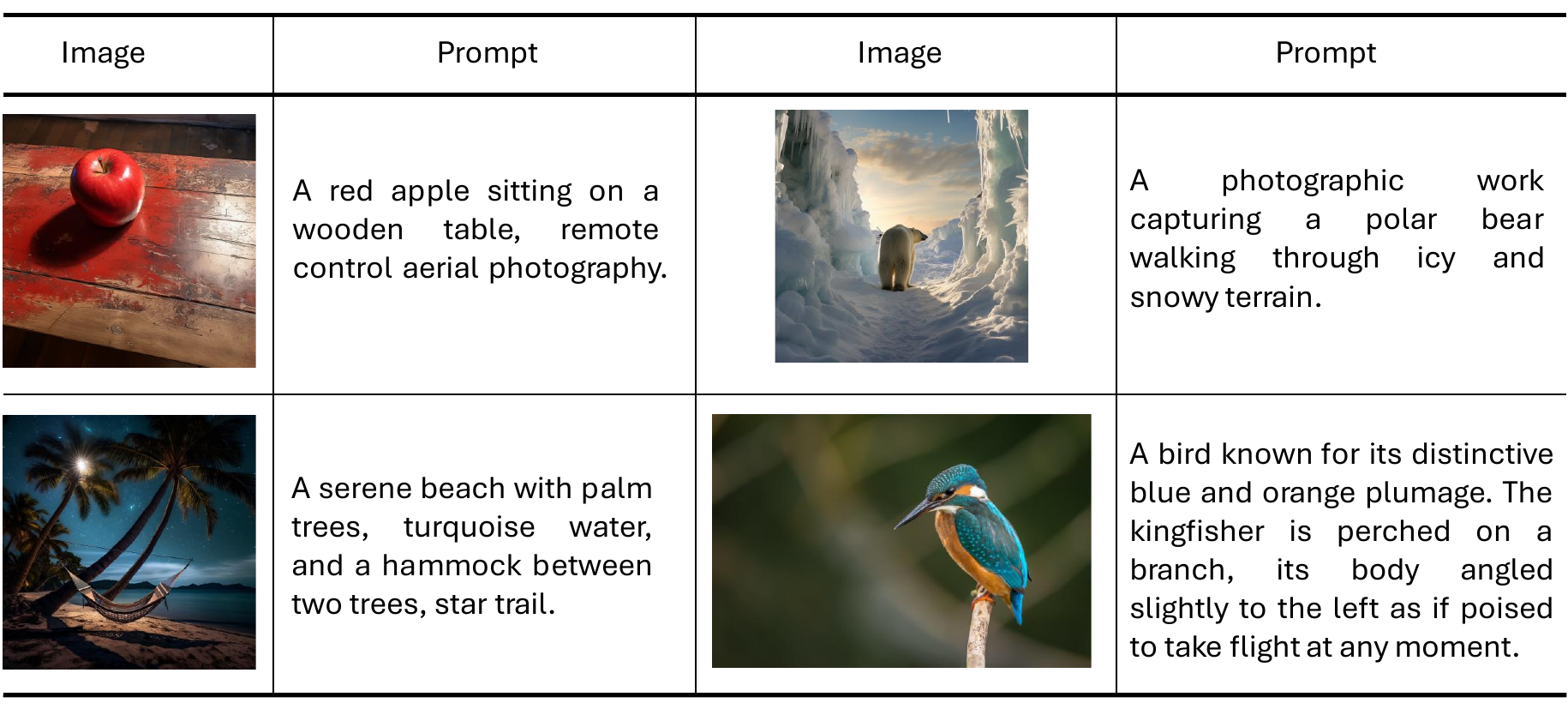}
\caption{\textbf{\footnotesize Samples in our proposed High-Quality Evaluation Dataset}. The evaluation dataset presented in this paper contains samples of superior visual quality compared to those in COCO-30K.
}
\label{fig:supp_hq_eval_sample}
\end{figure}

We observe the evaluations performed on the MSCOCO dataset could not adequately fully capture a model's proficiency in aesthetics and text-image alignment~\cite{rombach2022high,playground-v2}. Thus, we propose an evaluation dataset that consisting of 30,000 high-quality, aesthetically pleasing text-image pairs for a more thorough assessment of a model's ability to generate visually appealing images. The visualizations in Fig.~\ref{fig:supp_hq_eval_sample} exemplify the dataset's superior aesthetic quality and the precise alignment between textual descriptions and visual content.

\subsection{FID Comparisons with Open-source Models}
\begin{table}[t]
\centering
\caption{\textbf{Comparisons on FID and Clip-Score with Open-sourced T2I Models.} \model demonstrates enhanced performance in terms of FID and Clip-Score on the curated 30K High-Quality Evaluation Dataset.}\label{tab:supp-comparisons}
{
    \begin{tabular}{l c c c}
    \toprule
    Models     & $\mathrm{\#}$Params (B)  & FID $\downarrow$ & CLIP-Score $\uparrow$ \\
    \midrule
        Stable 1.5  & 0.9 & 17.03& 0.2748 \\
        Stable Turbo & 3.1  & 10.91& 0.2804 \\
        Stable XL  & 2.6& 7.38 & 0.2913 \\
        Stable Cascade & 5.1 & 9.96& 0.2839 \\
        Playground-V2.0 & 2.6& 8.68  & 0.2885\\
        Playground-V2.5 & 2.6 & 7.64 & 0.2871 \\
    \midrule
        \modelalpha  & 0.6 & 8.65 & 0.2787\\
        \model & 0.6 & 8.23 & 0.2797\\
    \bottomrule
    \end{tabular}
}
\vspace{-1em}
\end{table}

We conducted comparative analyses of open-source models using FID and CLIP-Scores on our curated dataset. The results, presented in Tab.~\ref{tab:supp-comparisons}, reveal that the weak-to-strong fine-tuning process significantly improves the model's ability to generate high-quality images and to align more closely with the given instructions, as \model shows a lower FID (8.65 \textit{v.s.} 8.23) and higher Clip-Scores (0.2787 \textit{v.s} 0.2797) compared to \modelalpha. Besides, compared to other models, the \model still shows comparable or even better performance with relatively small network parameters (0.6B).

\subsection{Designed Prompts for AI preference Study}

\begin{figure}[!ht]
\centering
\includegraphics[width=0.99\linewidth]{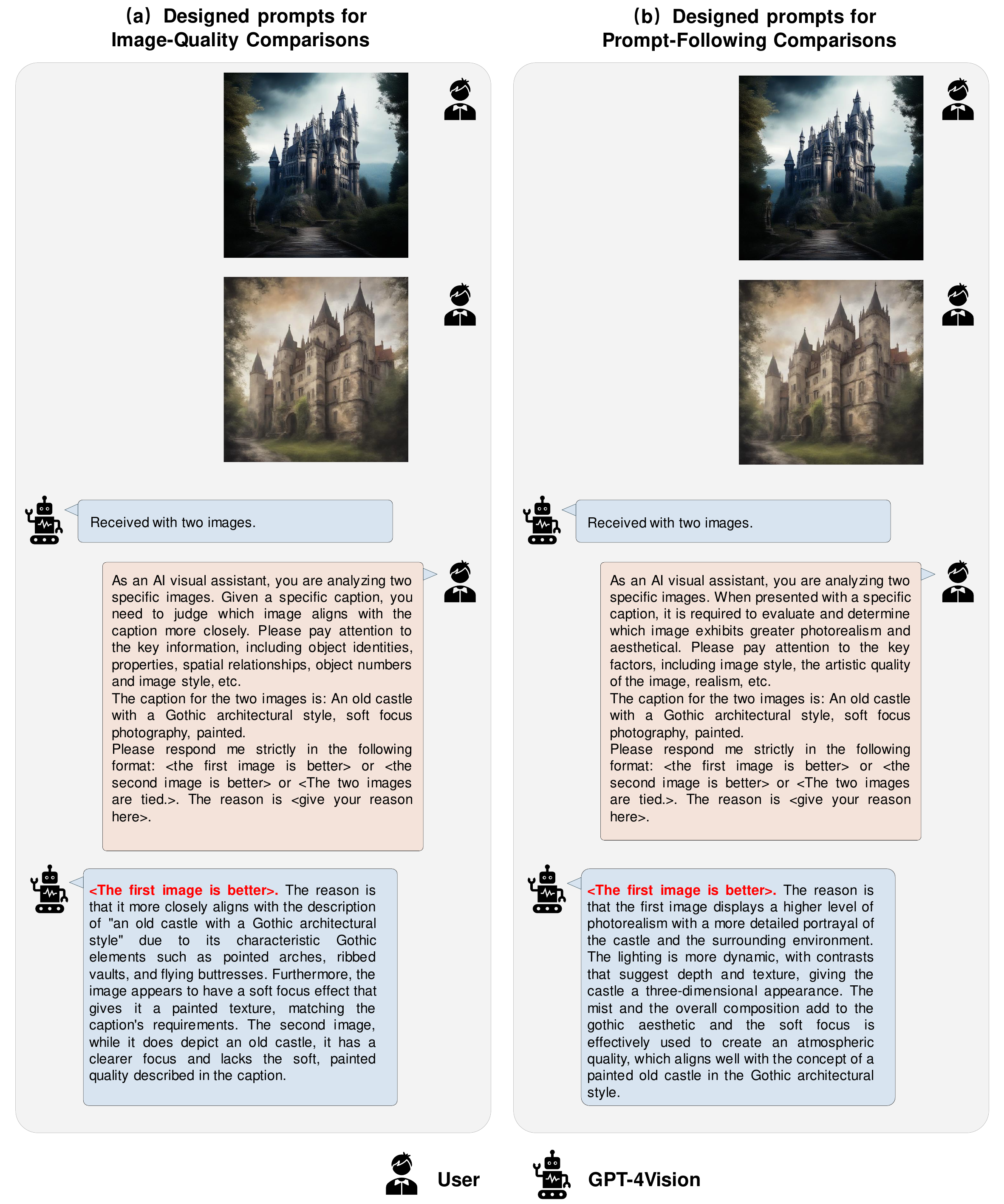}
\caption{\textbf{\footnotesize Illustration of the designed prompts for GPT-4Vision.}: We developed various prompts for GPT-4V to assess the image quality and instruction-following ability of different T2I generators, respectively.
\vspace{-1em}
}
\label{fig:supp_prompts}
\end{figure}

To study AI preferences among various text-to-image (T2I) generators, we employ the advanced multi-modality model, GPT-4V, as an automated evaluator. The instructions given to GPT-4V for comparing the generated images based on quality and adherence to prompts are illustrated in Fig.~\ref{fig:supp_prompts}. GPT-4V demonstrates the ability to provide logical assessments and reasons that coincide with human preferences.

\subsection{Pseudo-code for KV Token Compression}
We present the PyTorch-style pseudo-code for the KV Token Compression algorithm described in Sec.3.2 in the main paper. The implementation is straightforward.

\begin{algorithm}[!ht]
\small
\caption{\small KV Token Compression.}
\label{alg:attention}
\definecolor{codeblue}{rgb}{0.25,0.5,0.5}
\definecolor{codekw}{rgb}{0.85, 0.18, 0.50}
\lstset{
  backgroundcolor=\color{white},
  basicstyle=\fontsize{7.5pt}{7.5pt}\ttfamily\selectfont,
  columns=fullflexible,
  breaklines=true,
  captionpos=b,
  commentstyle=\fontsize{7.5pt}{7.5pt}\color{codeblue},
  keywordstyle=\fontsize{7.5pt}{7.5pt}\color{codekw},
  escapechar={|}, 
}
\begin{lstlisting}[language=python]
import torch
import torch.nn as nn
import xformer

class AttentionKVCompress(nn.Module):
  def __init__(self, dim, sampling='conv', sr_ratio=1,, **kwargs):
    super().__init__()
    
    # Projection layers and non-relavent definitions are omitted.
    self.sampling=sampling    # [`conv', `ave', `uniform']
    self.sr_ratio = sr_ratio
    if sr_ratio > 1 and sampling == "conv":
        # Avg Conv Init.
        self.sr = nn.Conv2d(dim, dim, groups=dim, kernel_size=sr_ratio, stride=sr_ratio)
        self.sr.weight.data.fill_(1/sr_ratio**2)
        self.sr.bias.data.zero_()
        self.norm = nn.LayerNorm(dim)
    
    def downsample_2d(self, tensor, H, W, scale_factor, sampling=None):
        B, N, C = tensor.shape
        tensor = tensor.reshape(B, H, W, C).permute(0, 3, 1, 2)

        new_H, new_W = int(H / scale_factor), int(W / scale_factor)
        new_N = new_H * new_W

        if sampling == "ave":
            tensor = F.interpolate(
                tensor, scale_factor=1 / scale_factor, mode='nearest'
            ).permute(0, 2, 3, 1)
        elif sampling == "uniform":
            tensor = tensor[:, :, ::scale_factor, ::scale_factor].permute(0, 2, 3, 1)
        elif sampling == "conv":
            tensor = self.sr(tensor).reshape(B, C, -1).permute(0, 2, 1)
            tensor = self.norm(tensor)

        return tensor.reshape(B, new_N, C).contiguous(), new_N
        
    def forward(self, x, mask=None, HW=None, block_id=None, ):
        B, N, C = x.shape
        new_N = N
        H, W = HW
        qkv = self.qkv(x).reshape(B, N, 3, C)
        q, k, v = qkv.unbind(2)
        if self.sr_ratio > 1:
            k, new_N = self.downsample_2d(k, H, W, self.sr_ratio, sampling=self.sampling)
            v, new_N = self.downsample_2d(v, H, W, self.sr_ratio, sampling=self.sampling)
        
        x = xformers.ops.memory_efficient_attention(q, k, v)
        
        x = x.view(B, N, C)
        x = self.proj(x)
        x = self.proj_drop(x)
        return x
    
\end{lstlisting}
\end{algorithm}

\subsection{\model \vs T2I products}
\begin{figure}[!ht]
\centering
\footnotesize
\includegraphics[width=0.99\linewidth]{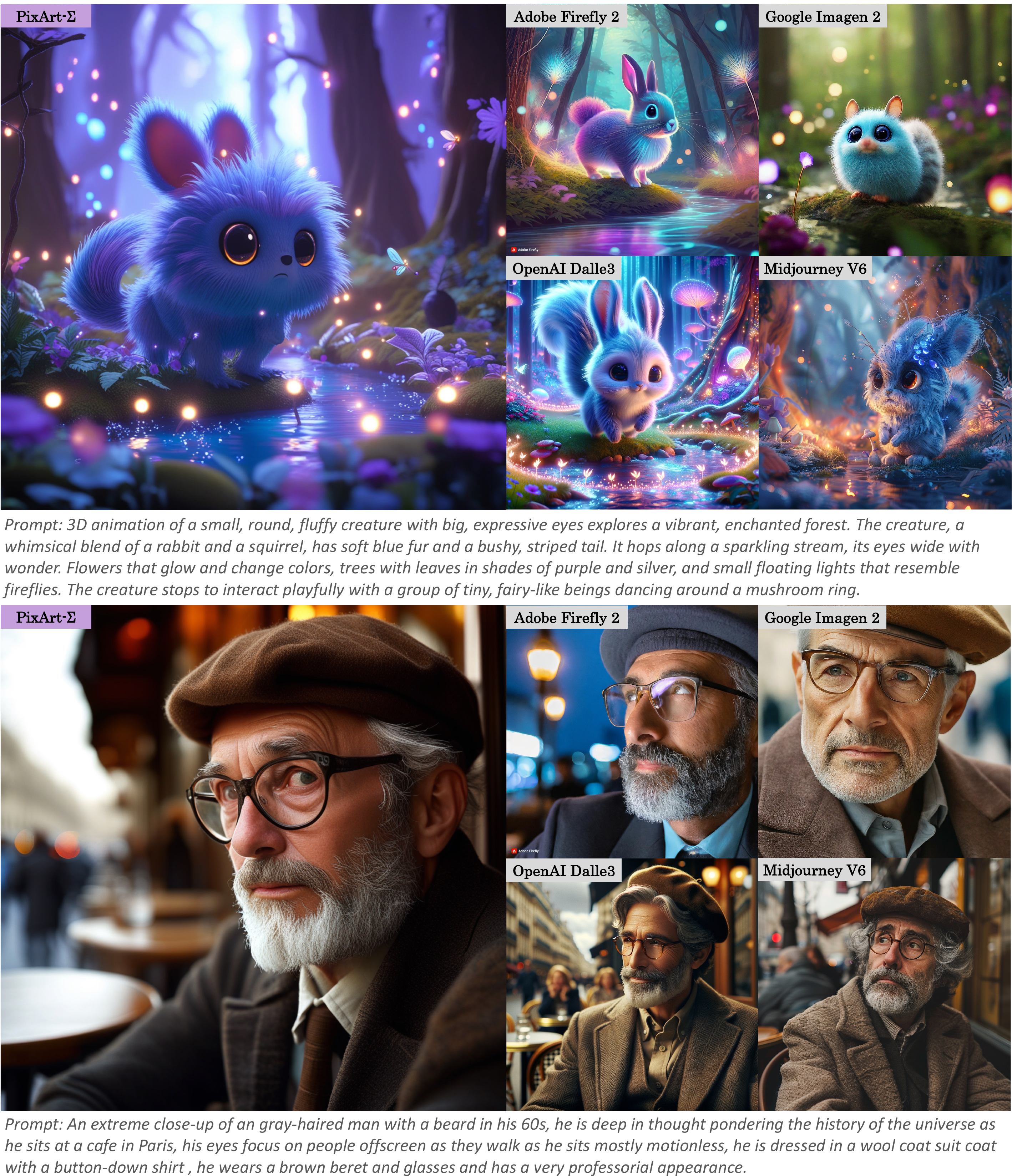}
\caption{\textbf{Compare \model and four other T2I products}: Firefly 2, Imagen 2, Dalle 3, and Midjourney 6. Images generated by \model are very competitive with these commercial products.}
\label{fig:compare_sota_supp}
\end{figure}

\begin{figure}[!ht]
\centering
\footnotesize
\includegraphics[width=0.88\linewidth]{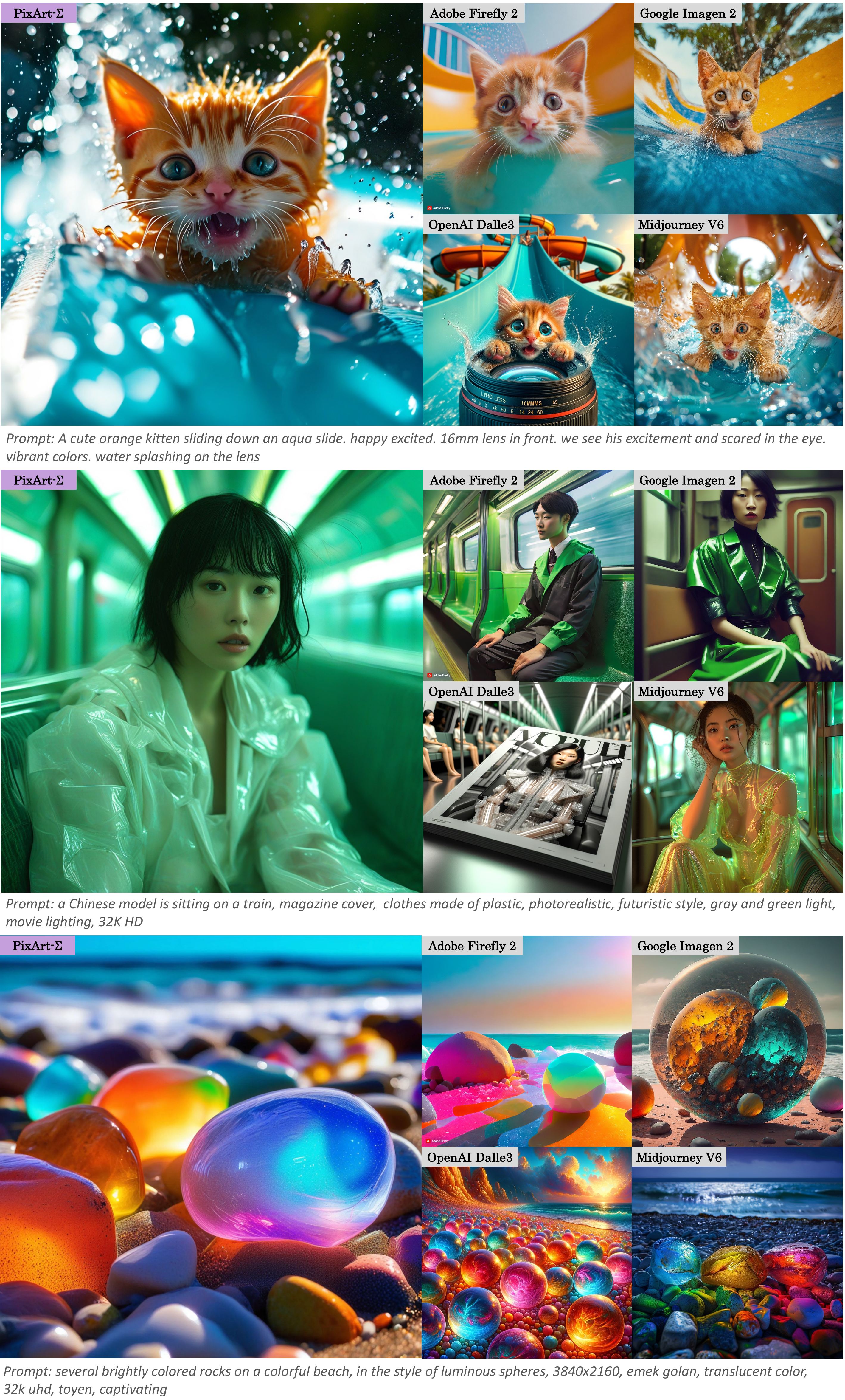}
\vspace{-1em}
\caption{\textbf{Compare \model and four other T2I products}: Firefly 2, Imagen 2, Dalle 3, and Midjourney 6. Images generated by \model are very competitive with these commercial products.}
\vspace{-2em}
\label{fig:compare_sota_supp2}
\end{figure}

We compare \model with four other close-source T2I products in Fig.~\ref{fig:compare_sota_supp}, and Fig.~\ref{fig:compare_sota_supp2}. Our model can produce high-quality, photo-realistic images with rich details and is comparable with these products.

\subsection{More images generated by \model}
\begin{figure}[!ht]
\centering
\footnotesize
\includegraphics[width=0.95\linewidth]{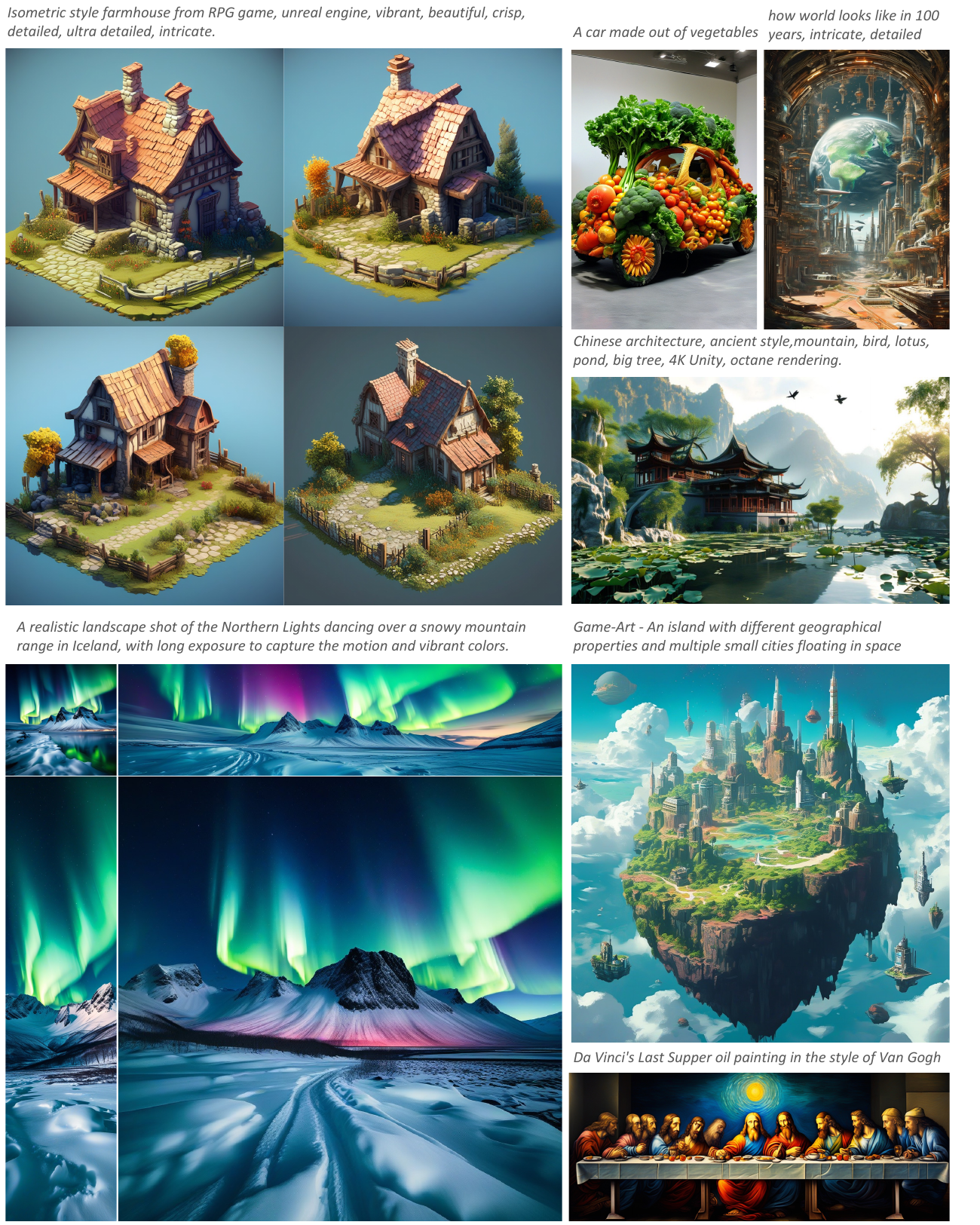}
\caption{\textbf{Illustrations of High-quality images generated by \model.} \model can generate high-quality images with fine-grained details, and diverse images with different aspect ratios.}
\label{fig:omega_images_supp}
\end{figure}

\begin{figure}[!ht]
\centering
\footnotesize
\includegraphics[width=0.90\linewidth]{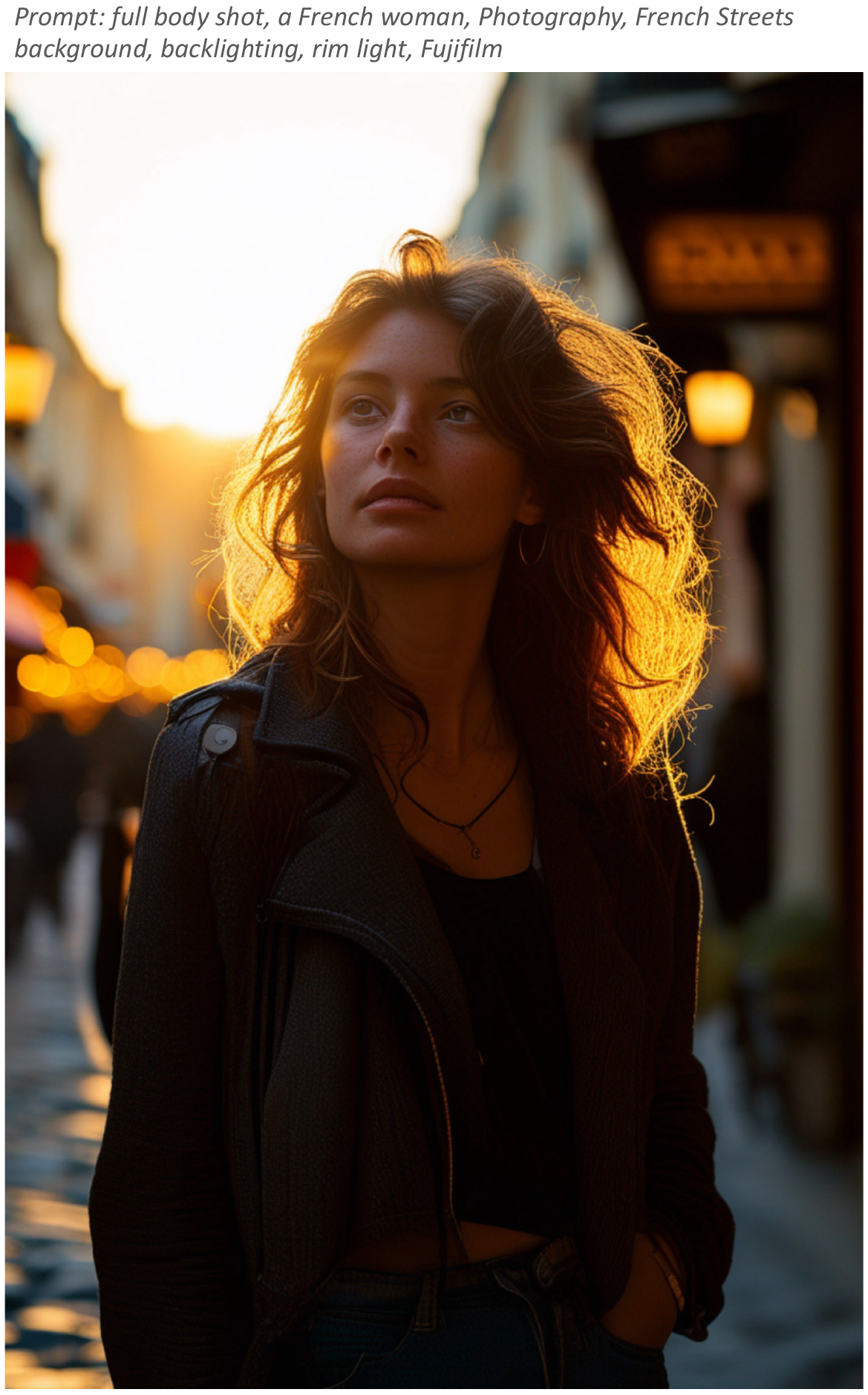}
\vspace{-1em}
\caption{\textbf{High-resolution (4K) images generated by \model.} \model can directly generate high-quality 4K HD (3840$\times$2560) images while preserving fine-grained details.}
\vspace{-1em}
\label{fig:omega_images_supp2}
\end{figure}

\begin{figure}[!ht]
\centering
\footnotesize
\includegraphics[width=0.90\linewidth]{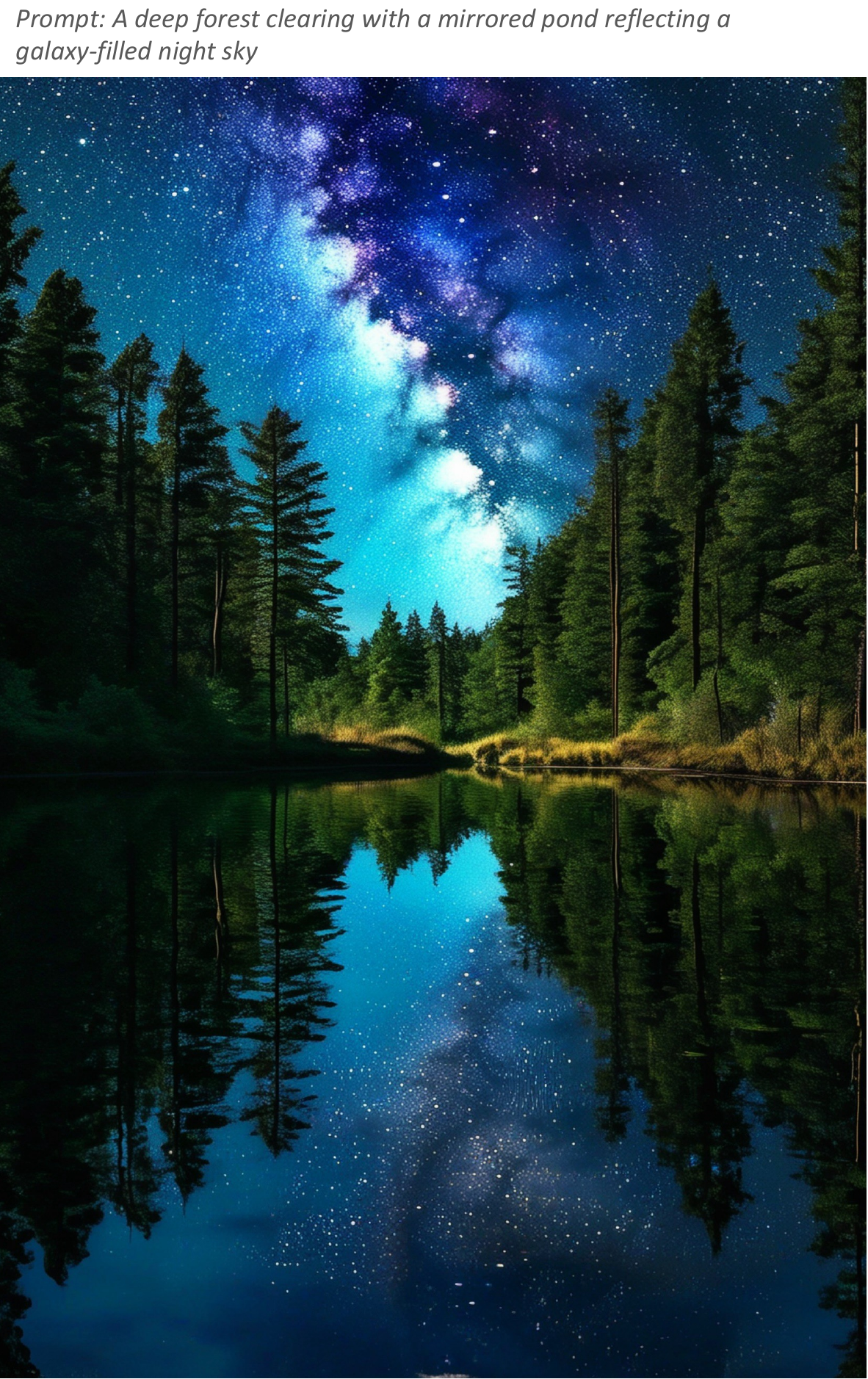}
\vspace{-1em}
\caption{\textbf{ High-resolution (4K)  images generated by \model.} \model can directly generate high-quality 4K HD (3840$\times$2560) images while preserving fine-grained details.}
\vspace{-1em}
\label{fig:omega_images_supp1}
\end{figure}

Fig.~\ref{fig:omega_images_supp}, ~\ref{fig:omega_images_supp2} and ~\ref{fig:omega_images_supp1} showcase additional visual outputs produced by \model. The quality of these samples is remarkable, characterized by their high fidelity and accuracy in closely matching the provided textual prompts.

\subsection{Limitation and Social Impact}
\noindent \textbf{Limitation.}
Our model still lacks the ability to generate some specific scenes and objects, especially text generation and hand generation. It is not perfect in the following aspects: it cannot fully align the complex prompts input by the user, face generation may have flaws and sensitive content may be generated. Subsequent research should focus on higher-quality data construction, scale model size, and improving model illusion and security issues through super alignment.

\noindent\textbf{Negative social impact.}
Text-to-image models may bring a negative social impact by generating images that present stereotypes or discriminate against certain groups. For instance, the images generated by a text-to-image model may depict unbalanced proportions of gender and inaccurate content for some uncommonly used concepts. Mitigating these issues requires careful data collection.


\clearpage
\bibliographystyle{splncs04}
\bibliography{main}
\end{document}